\pdfoutput=1

\documentclass[11pt]{article}

\usepackage{acl}

\usepackage{times}
\usepackage{latexsym}
\usepackage{graphicx}
\usepackage[normalem]{ulem}

\definecolor{lightgray}{gray}{0.95}

\usepackage[T1]{fontenc}

\usepackage[utf8]{inputenc}

\usepackage{microtype}
\usepackage{multirow}
\usepackage{amsfonts,amsmath,bm}

\title{Multilingual Summarization with Factual~Consistency~Evaluation}

\author{
  Roee Aharoni\thanks{\hspace{5px}Equal contribution.} \\ 
  Google Research \\ {\small\tt roeeaharoni@google.com} \And
  Shashi Narayan$^{\ast}$ \\ Google DeepMind \\ {\small\tt shashinarayan@google.com} \And
  Joshua Maynez \\ Google DeepMind \\ {\small\tt joshuahm@google.com}  \AND
  \textbf{Jonathan Herzig} \\  Google Research \\ {\small\tt jherzig@google.com} \And 
  \textbf{Elizabeth Clark} \\ Google DeepMind \\ {\small\tt eclark@google.com} \And
  \textbf{Mirella Lapata}\\ Google DeepMind \\ {\small\tt lapata@google.com}}

\begin{document}
\maketitle
\begin{abstract}
  Abstractive summarization has enjoyed renewed interest in recent
  years, thanks to pre-trained language models and the availability of
  large-scale datasets. Despite promising results, current models
  still suffer from generating factually inconsistent summaries,
  reducing their utility for real-world application. Several recent
  efforts attempt to address this by devising models that
  automatically detect factual inconsistencies in machine generated
  summaries. However, they focus exclusively on English, a language
  with abundant resources. In this work, we leverage factual
  consistency evaluation models to improve \emph{multilingual}
  summarization. We explore two intuitive approaches to mitigate
  hallucinations based on the signal provided by a multilingual NLI
  model, namely data filtering and controlled generation. Experimental
  results in the 45 
languages from the XLSum dataset show gains over
  strong baselines in both automatic and human evaluation. 
  We release models and human judgements of summaries to foster progress towards more factually consistent multilingual
  summarization.\footnote{\url{https://github.com/google-research/google-research/tree/master/mface}}
\end{abstract}

\section{Introduction}

The past few years have witnessed a huge leap forward in abstractive
summarization thanks to large-scale pretraining
\cite{devlin-etal-2019-bert,lewis-etal-2020-bart} and the availability
of benchmark datasets.  A well-known issue limiting the wider adoption
of abstractive summarization models is their tendency to generate
factually inconsistent summaries, a.k.a ``hallucinations''
\cite[][\emph{inter alia}]{maynez-etal-2020-faithfulness,
  zhao-etal-2020-reducing}. A recently popular line of work explores
how to best detect hallucinations in machine generated text, thereby
enabling the automatic identification of factually inconsistent
summaries \cite[][\emph{inter
  alia}]{eyal-etal-2019-question,falke-etal-2019-ranking,kryscinski-etal-2020-evaluating,
  wang-etal-2020-asking,goyal-durrett-2021-annotating,scialom-etal-2021-questeval,honovich-etal-2022-true,Tang:ea:2022}.

While such approaches may prove useful for automatic evaluation, it
remains unclear how to best leverage them for \textit{improving}
summarization models \emph{in multiple languages}. While focusing
exclusively on English, previous work suggests different techniques to
this effect such as discarding ``noisy'' training examples
\cite{gehrmann-etal-2021-gem}, contrastive learning paradigms
\cite{nan-etal-2021-improving}, controlled generation and planning
\cite{narayan-etal-2021-planning,rashkin-etal-2021-increasing}, or
reinforcement-learning approaches that use the evaluation model score
as a reward function
\cite{gunasekara-etal-2021-using-question}. Despite promising results,
no method has emerged as a clear winner in English, let alone across
languages with varying amounts of data and resources.

\begin{figure}[!t]
    \centering
    \frame{\includegraphics[scale=0.11]{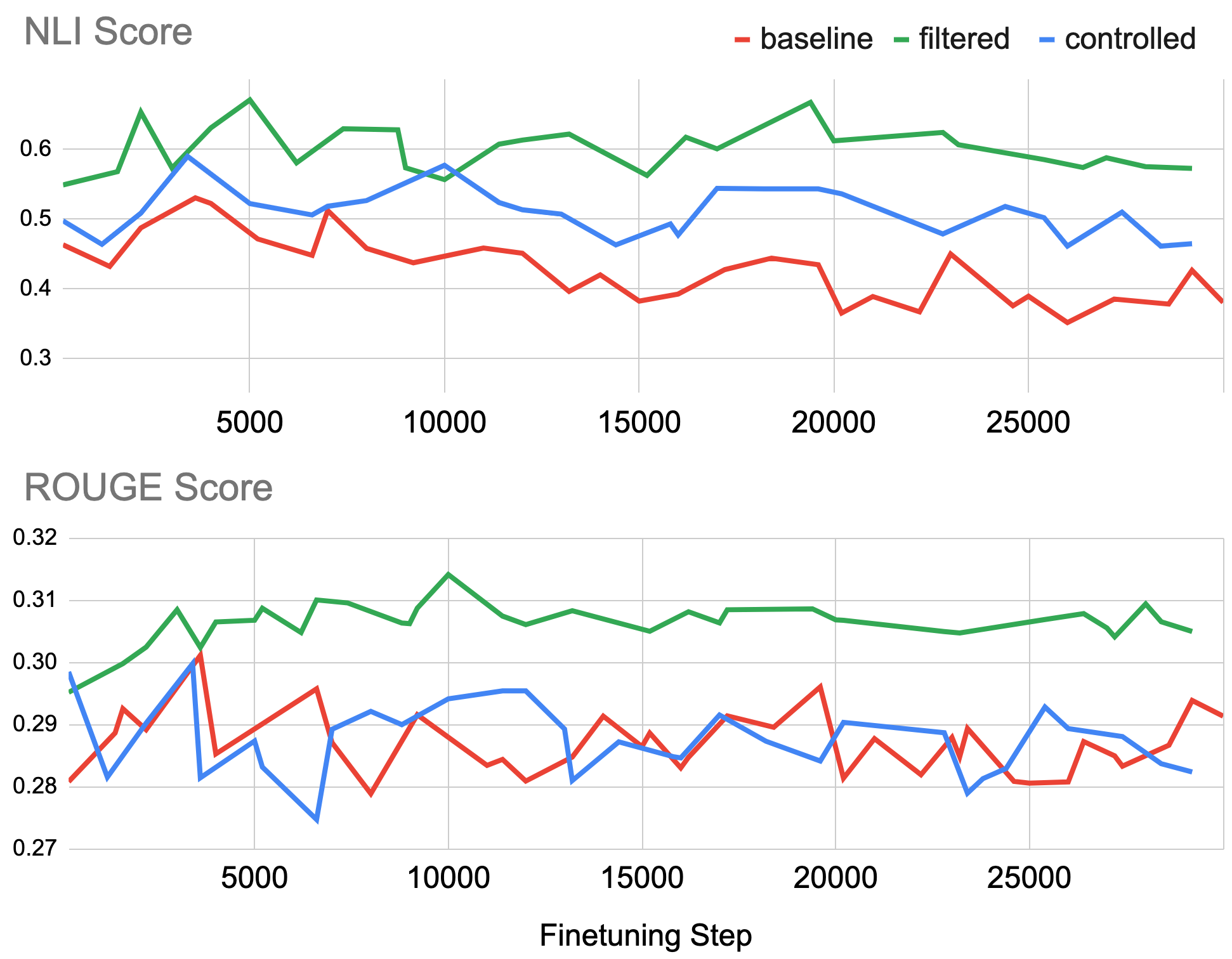}}
    \caption{NLI and ROUGE scores for different models on the Arabic
      development set of XLSUM during fine-tuning. Using a multilingual entailment model during training (via data filtering or controlled generation) improves summary quality over a baseline model trained without using the entailment signal.}
    
    \label{fig:scores_arabic}
    \vspace{-10pt}
\end{figure}

In this work, we  leverage factual consistency evaluation
models to improve summarization systems in multiple
languages. Specifically, we employ Textual Entailment models
(a.k.a. Natural Language Inference;  \citealt{dagan2005pascal,
  bowman2015large}) in order to determine whether a summary is
factually consistent
\cite{maynez-etal-2020-faithfulness,laban2021summac}. We
opportunistically opt for NLI given the availability of multilingual
benchmarks for model training
\cite{conneau-etal-2018-xnli,nie-etal-2020-adversarial}. Approaches
based on question generation and answering have been also shown to
work well for factuality evaluation
\cite[i.e.][]{scialom-etal-2021-questeval, honovich-etal-2021-q2,
  deutsch-etal-2021-towards}, however, they are not easily portable
due to the scarcity of respective resources in languages other than
English.

We first analyze the quality of the training data for summarization
models using a strong multilingual NLI model (as evaluated on the XNLI dataset, 
\citeauthor{conneau-etal-2018-xnli}; \citeyear{conneau-etal-2018-xnli}). In particular, we
train our multilingual NLI model, following the guidelines from the TRUE survey \cite{honovich-etal-2022-true} for the assessment of factual
consistency.
Focusing on the XLSum\footnote{Available for non-commercial use, see: \url{http://github.com/csebuetnlp/xl-sum\#license}} multilingual summarization dataset
\cite{hasan-etal-2021-xl}, we find that for some languages up to
70\%~of training examples are not factually consistent according to
the NLI model, while such examples are commonly used for training. We
use the NLI signal to improve the quality of the generated summaries
in two ways: (1)~{\em data filtering}, where we only train on examples
whose summaries are predicted to be entailed by the input, and (2)
{\em controlled generation}, where we also leverage ``negative''
training examples by conditioning the summarization model on the NLI
signal.
We evaluate the proposed approaches using both automatic and human evaluation in 45~languages, and observe significant
gains in the faithfulness of the generated summaries over strong baselines. Finally, we show that the human judgments we collected in all languages are useful for training automatic metrics to assess the quality, factual consistency and informativeness of generated summaries.


To summarize, the contributions of this work are three-fold: (1) we
analyze the quality of the \mbox{XLSum} dataset
\cite{hasan-etal-2021-xl} using strong multilingual NLI models and
reveal severe issues with faithfulness in the training data across
languages; (2)~we explore methods for improving downstream
summarization models trained on this data using a multilingual NLI
signal, and show large gains in both automatic and human evaluation;
and (3)~using the data from our large-scale human evaluation study, we learn metrics for automatically evaluating summaries in multiple languages along the dimensions of Quality, Factual Consistency, and Informativeness.\footnote{We will release our NLI models, summarization models, and human
judgments of summaries in multiple languages to foster future work on
this task.} To the best of our knowledge, our work is the first to
examine the faithfulness of summarization systems in  multilingual
settings, and we hope it will encourage the development of better
metrics and models in multilingual text generation.


\section{Related Work}

Despite significant improvements in recent  years
\cite{liu-lapata-2019-text,lewis-etal-2020-bart,Raffel:ea:2019}, 
abstractive summarization models are still prone to ``hallucination'',
i.e., the inclusion of factual errors in the generated summaries
\cite{song-etal-2018-structure,maynez-etal-2020-faithfulness,kryscinski-etal-2020-evaluating,gabriel-etal-2021-go}. 

A plethora of approaches have been proposed for the automatic
detection of factual inconsistencies in machine generated text (see
\citealt{honovich-etal-2022-true} and \citealt{Tang:ea:2022} for
overviews) with varying degrees of success. There is growing consensus
that techniques based on textual entailment
\cite{maynez-etal-2020-faithfulness,goyal-etal-2021-multi,goyal-durrett-2021-annotating}
and question generation and answering models
\cite{durmus-etal-2020-feqa,wang-etal-2020-asking,deutsch-etal-2021-towards,Fabbri:ea:2021,scialom-etal-2021-questeval,honovich-etal-2021-q2}
achieve strong performance across tasks and datasets
\cite{laban2021summac,honovich-etal-2022-true}.  Another line of
 work uses synthetically generated data to train
models for evaluating factual consistency
\cite{kryscinski-etal-2020-evaluating,zhao-etal-2020-reducing,goyal-durrett-2020-evaluating}.

Aside from assessing system output, several studies have proposed
novel model architectures which enforce factuality during training or
inference. These include extracting facts from the source and
incorporating them as additional input to the model
\cite{Cao:ea:2018,aralikatte-etal-2021-focus,zhu-etal-2021-enhancing}, planning using entity chains and avoiding entities that are not in the input \cite{narayan-etal-2021-planning,narayan-etal-2022-well},
using reinforcement learning to optimize model training with factual
correctness as a reward
\cite{zhang-etal-2020-optimizing,arumae-liu-2019-guiding,pasunuru-bansal-2018-multi,nan-etal-2021-improving},
reranking candidate summaries within a beam using entailment
predictions \cite{falke-etal-2019-ranking} or quantity verification
scores \cite{zhao-etal-2020-reducing}, using contrastive learning
\cite{cao-wang-2021-cliff,wan-bansal-2022-factpegasus}, modifying the
training objective to only maximize the likelihood of factual words
\cite{goyal-durrett-2021-annotating}, incorporating factuality into the
pretraining objective of models tailored to text summarization tasks
\cite{wan-bansal-2022-factpegasus}, and adaptively removing examples
with high log loss \cite{kang-hashimoto-2020-improved}. Other work
simply removes noisy training samples
\cite{nan-etal-2021-entity,goyal-durrett-2021-annotating} in the hope
that factuality will improve by training on better examples.

Despite promising results, it is unclear whether previous techniques
transfer to languages beyond English.  Our own work aims to improve
the factuality of abstractive summarization \textit{across
  languages}. Leveraging recent progress on multilingual pretrained
models \cite{xue-etal-2021-mt5}, we show that entailment-based metrics
can be trained to detect factually-inconsistent summaries in multiple languages,
and that this signal can be leveraged to improve summarization systems
in those languages.



\section{Multilingual Factual Consistency Evaluation}
\label{sec:multnli}



We cast factual consistency evaluation as a Natural Language Inference
(NLI) task. The input forms the premise, the summary forms the
hypothesis~\cite{maynez-etal-2020-faithfulness,laban2021summac,honovich-etal-2022-true},
and the NLI model is used to predict whether the summary is entailed
by the input. More formally, given input document~$d$ and summary~$s$,
we define an NLI model~$\mathcal{M}$ as a binary classifier, where
$\mathcal{M}(d, s) \approx p(s~\textrm{is entailed by}~d)$.

Recent studies \cite{honovich-etal-2022-true} on evaluating factual consistency in summarization and
other related tasks in English have
obtained promising results when finetuning large pretrained models on
NLI datasets. Specifically, they finetune the~T5 pretrained
encoder-decoder models \cite{2020t5} for binary classification where
the entailment relation translates to a positive label and
contradiction/neutral relations are merged to a negative label. Their
model encodes the concatenation of the premise (document) and
hypothesis (summary) and decodes a single token that represents the
class label (entailment or no entailment).\footnote{Other work
  \cite{laban2021summac, schuster2022stretching} breaks documents into
  sentences before running NLI models, however, we refrained from
  doing so to avoid loss of context.}

Since we are interested in  evaluating factual consistency in multiple
languages,     we     extend      the     modeling     approach     of
\citet{honovich-etal-2022-true}  to a  multilingual  setting.  As  our
pretrained  model, we  use mT5-XXL~\cite{xue-etal-2021-mt5}  which was
trained  on  mC4,  a  dataset  drawn  from  the  public  Common  Crawl
covering~101    languages.    We    finetuned    mT5-XXL    on    the
ANLI~\cite{nie-etal-2020-adversarial}                              and
XNLI~\cite{conneau-etal-2018-xnli}   datasets.    ANLI   contains~162K
English-only examples, while XNLI has~37K examples\footnote{We use the
  XNLI development set for training as it was manually curated, unlike
  the XNLI  training set which  was automatically translated.}   in 15
languages.  As mentioned above for  the English case, the multilingual
model  is  trained   to  generate  a  binary  label   when  given  the
concatenation of  a premise and  hypothesis, where the  positive label
corresponds to an  entailment relation, and the  negative label stands
for  a neutral/contradiction  relation. During  inference, we  score a
premise and hypothesis input by  measuring the output probability when
force-decoding the positive label, resulting  in a score between 0 (no
entailment) and 1 (entailment).

We measured the quality of our multilingual NLI model by evaluating on
the XNLI \cite{conneau-etal-2018-xnli} test set and the TRUE
benchmark~\cite{honovich-etal-2022-true}. The latter is a standardized
collection of datasets representing various tasks
(summarization, dialog generation, paraphrasing and
fact-checking) with manual annotations for factual consistency.  On
XNLI, our model yields an average accuracy of~90.0 over 15 languages
in comparison to 87.8 reported in
\citet{xue-etal-2021-mt5}.\footnote{The numbers are not fully
  comparable as \citet{xue-etal-2021-mt5} classify into three classes
  while we use binary classification.} We present results for
individual languages in Appendix~\ref{sec:appendix_nli}.  On the
(English-only) TRUE benchmark, our model's average ROC AUC is~82.4 in
comparison to~83.4 reported in~\citet{honovich-etal-2022-true} for
their best performing English-only, T5-11B
model~\cite{2020t5} trained on ANLI. While our model is trained on both ANLI (English) and
XNLI (15~languages, detailed in Table~\ref{tab:training_data}), we
assume it can generalize to additional languages (for which NLI data
is not available) due to the nature of the pretrained model (mT5,
trained on 101 languages).

\begin{table}[t]
\begin{center}
\resizebox{1\columnwidth}{!}{
  \begin{tabular}{@{}l@{~}r@{~}c@{~}|l@{~}r@{~}c@{}} 
\multicolumn{1}{@{}c@{}}{Language}  & \multicolumn{1}{c@{~}}{\# Train} &
\multicolumn{1}{c@{~}|}{Ent\%} & \multicolumn{1}{c@{}}{Language}  &
\multicolumn{1}{c@{}}{\# Train}  & \multicolumn{1}{c@{}}{Ent\%} \\ \hline
Amharic            & 5,761                    & 42.67       &                      Pidgin             & 9,208                       & 39.09 \\                                                          
\uline{Arabic}             & 37,519                        & 45.19         &          Portuguese         & 57,402                        & 40.19         \\                                                  
Azerbaijani        & 6,478                          & 34.15         &                       Punjabi            & 8,215                    & 28.29         \\                                                  
Bengali            & 8,102                    & 48.58         &                       \uline{Russian}           & 62,243                 & 40.50         \\                                      
Burmese            & 4,569                          & 32.48         &                       Scottish (G)    & 1,313                        & 42.42         \\                                                   
\uline{Chinese (S)}    & 37,362                    & 50.81         &          Serbian (C) & 7,275                          & 31.08         \\                                                  
\uline{Chinese (T)}   & 37,373                     & 50.59         &           Serbian (L)    & 7,276                         & 30.61         \\                                                  
\uline{English}            & 306,522              & 60.00         &          Sinhala            & 3,249                       & 36.53         \\                                                  
\uline{French}             & 8,697                  & 36.96         &          Somali             & 5,962                    & 38.49         \\                                                  
Gujarati           & 9,119                          & 29.60         &                       \uline{Spanish}            & 38,110                    & 35.29         \\                                     
Hausa              & 6,418                          & 39.42         &                       \uline{Swahili}            & 7,898                      & 42.83         \\                                     
\uline{Hindi}              & 70,778                & 43.68         &          Tamil              & 16,222                      & 50.63         \\                                                  
Igbo               & 4,183                          & 35.64         &                       Telugu             & 10,421                  & 35.19         \\                                                  
Indonesian         & 38,242                        & 52.48         &                       \uline{Thai}               & 6,616              & 46.84         \\                                     
Japanese           & 7,113                          & 68.96         &                       Tigrinya           & 5,451                      & 36.93         \\                                                  
Kirundi            & 5,746                          & 44.27         &                       \uline{Turkish}            & 27,176                & 44.31         \\                                     
Korean             & 4,407                          & 48.92         &                       Ukrainian          & 43,201                     & 38.27         \\                                                  
Kyrgyz             & 2,266                           & 32.48         &                        \uline{Urdu}               & 67,665               & 42.17         \\                                     
Marathi            & 10,903                         & 29.28         &                       Uzbek              & 4,728                       & 36.40         \\                                                  
Nepali             & 5,808                         & 51.58         &                       \uline{Vietnamese}          & 32,111               & 36.12         \\                                    
Oromo              & 6,063                          & 40.76         &                       Welsh              & 9,732                       & 49.38         \\                                                  
Pashto             & 14,353                         & 44.83         &                       Yoruba             & 6,350                        & 37.15         \\                                                  
Persian            & 47,251                        & 49.19         &
Avg            & 24,670                      & 41.37
\\   \hline 
\end{tabular}
} 
\end{center}
\vspace{-10px}
\caption{Statistics on  XLSum training data: total
  number of examples per language, proporion of examples where
  the summary was entailed by the input (\%~Ent). Languages   in
  XLSum \emph{and} XNLI are \uline{underlined}, for other lanuages NLI
  classification is zero-shot. Chinese (S/T) refers to
  simplified/traditional; Serbian (C/L) is a shorthand for Cyrilic and
Latin respectively; and Scottish (G) abbreviates Gaelic. }
\label{tab:training_data}
\vspace{-20px}
\end{table}

\section{Summarization Models}

We next describe two summarization approaches which exploit the
factual consistency evaluation signal provided by the multilingual NLI model. 

\subsection{Data Filtering}

An intuitive approach to improving the factuality of machine generated
summaries is to enhance the quality of the training data, simply by
filtering noisy training samples
\cite{nan-etal-2021-entity,goyal-durrett-2021-annotating}.  More formally, given a training corpus~$\mathbb{D}$ of input document-summary pairs, we find ~$\mathbb{D^+} \subset \mathbb{D}$ such that for each document-summary pair $(d, s) \in \mathbb{D^+}$, \mbox{$p(s~\textrm{is entailed by}~d) > 0.5$}.

We used our multilingual NLI model (see Section~\ref{sec:multnli}) to
annotate the training data in XLSum \cite{hasan-etal-2021-xl} for all
45~languages. Table~\ref{tab:training_data} shows the total number
training examples and the proportion where the summary was predicted
to be entailed by the input (using a threshold of 0.5 on the NLI model
score). The proportion of entailed summaries ranges from~68.96\% (for
Japanese) to~28.29\% (for Punjabi). For all but three languages
(English, Japanese, Nepali), the NLI model predicted \textit{less than
  half} of the training summaries as being entailed by the input. We
find these numbers strikingly low; this may be due to the nature of
the dataset, since the relationship between news headlines and their
corresponding article can be somewhat loose (e.g.,~headlines may
include ``clickbate'' and additional details that are not mentioned in
the article). Another reason might be errors of the NLI model; while
it was shown to work well on TRUE/XNLI, XLSum may represent a
different distribution.

Overall, the results in Table~\ref{tab:training_data} indicate that
filtering the training data based on the NLI signal can have a large
impact on the resulting summarization model.  Training on the entailed
portion of the data may result in more factual summaries, however, at
the expense of summary quality as the model unavoidably sees fewer
examples (e.g.,~there are only 557~instances for Scottish Gaelic after
filtering, while the original training set has~1,313).



\subsection{Controlled Generation}
\label{sec:controlled}

Another way to leverage the NLI signal for improving the summarization
model is via controlled generation \cite{Keskar2019CTRLAC,
  rashkin-etal-2021-increasing}. In this approach, special tokens are
prepended to the model's input to indicate/control whether the output
should be entailed or not.

Let~$\mathbb{D}$ denote a training corpus of document~summary pairs
$(d,s)$. We annotate each $(d, s) \in \mathbb{D}$ as $(d', s)$, where
$d'$ is $d$ prepeneded with an ``\mbox{<entailed>}'' symbol if $p(s~\textrm{is entailed by}~d) >
0.5$, and otherwise $d'$ is $d$ prepended with ``\mbox{<not-entailed>}''.  The model trained
on~$\mathbb{D}$ enhanced with these annotations is expected to learn
the correlation between entailment and the special token value, and as
a result to be ``controlled'' to produce more faithful summaries by
prepending the token that corresponds to faithful (aka entailed)
summaries in inference time. This method \emph{implicitly} teaches the
model to learn from the entailment signal while taking advantage of
all available training data. It may, however, be more sensitive to
wrong predictions by the entailment model as noisy examples are not
discarded.


\section{Experimental Setup}
\label{sec:setup}

We focus on XLSum \cite{hasan-etal-2021-xl}, a recently
introduced summarization dataset. XLSum extends to multiple languages
the methodology put forward in \citet{narayan-etal-2018-dont} for the
creation of the English-only XSum; it contains 1~million BBC
article-summary pairs covering 45~languages.

\subsection{Model Details}
We finetuned three models based on mT5
XXL~\citep[][14B parameters]{xue-etal-2021-mt5}. The first is a ``Vanilla'' model which
is trained on the XLSum data as-is. As previous work has shown that
multilingual training improves performance for low-resource languages
\citep{aharoni-etal-2019-massively, hasan-etal-2021-xl}, we also
follow this setting and finetune a single massively multilingual model
for all 45~languages in XLSum. The second model (``Filtered'') is
finetuned only on the portion of the data that passed the multilingual
NLI filter. The third model (``Controlled'') is trained on all data,
using the controlled generation approach mentioned
above. Specifically, for control tokens ``\mbox{<entailed>}'' and ``\mbox{<not-entailed>}'', we used two extra spare tokens
from the mT5 vocabulary and prepended them to the input
\citep{Keskar2019CTRLAC,rashkin-etal-2021-increasing}. 
During inference, we always prepend the input with ``\mbox{<entailed>}'' and report on the whole development and test sets.

\begin{table}[t!]
\begin{center}
\begin{small}
\begin{tabular}{l|c|c}
\multicolumn{1}{c|}{Model}      & Best-NLI & \multicolumn{1}{c}{Best-ROUGE} \\
\hline
Vanilla    & 3,600     & 15,000      \\
Filtered   & 2,200     & 12,000      \\
Controlled & 3,400     & ~~8,800  \\ \hline
\end{tabular}
\end{small}
\end{center}
\vspace{-6pt}
\caption{Number of finetuning steps for  best checkpoint for each
  model according to NLI and ROUGE on the XLSum development set.}
\label{tab:checkpoints}
\vspace{-20px}
\end{table}
\

Ideally, we would like to evaluate a \emph{single} model checkpoint
for \emph{all} languages; in the literature,  the best checkpoint is
often selected using ROUGE. However, 
we also employ  NLI scores to quantify improvements in faithfulness. 
For each model, we select two checkpoints that are best according to
ROUGE and NLI (on the development set), when averaged across all
languages. Table~\ref{tab:checkpoints} summarizes the number of
finetuning steps that led to the best checkpoints for each model
according to ROUGE and NLI. For all models, we observe that best NLI
checkpoints are earlier than ROUGE-based ones.

\subsection{System Comparisons}
\label{subsec:comparisons}

We compare the above approaches to three additional
baselines. Firstly, we record the number of examples that pass the NLI
filter, per language, and select the same number at ``Random''. We
then finetune a model similarly to the ``Filtered'' model above using
this randomly selected data. Secondly, we introduce a ``Self-ROUGE''
baseline which selects examples where the ROUGE of the summary with
respect to the input document is highest. Again, we choose the same
number of examples as those which passed the NLI
filter, and finetune a model on this data. Finally, we compare
against model output from \citet{hasan-etal-2021-xl} who finetuned an
mT5-Base pretrained model.


\subsection{Automatic Evaluation}

We report ROUGE \cite{lin-2004-rouge} which is commonly used to
measure the informativeness and fluency of model  summaries
against gold-standard references.\footnote{For ROUGE, we used the python implementation from \url{https://github.com/google-research/google-research/tree/master/rouge}} We also quantify faithfulness with
the reference-free NLI score \cite[][\emph{inter
  alia}]{maynez-etal-2020-faithfulness,honovich-etal-2022-true}. Since there are no tokenizers available for many of the
languages in XLSum, we report ROUGE-L computed using the sentencepiece
tokenization of mT5. Regarding NLI, we compute for each summary
whether it is entailed by the input, and report the average over all
examples in a partition (test or development set).

\begin{figure}[t!]
    \centering
    \footnotesize
    \resizebox{0.85\columnwidth}{!}{
    \begin{tabular}{@{}p{7.7cm}@{}} \hline
    \textbf{Quality: Is the summary
        comprehensible?} \\\hline
    \textbf{Incomprehensible:} The summary is difficult to
    understand. It has serious grammatical errors, low fluency, and/or repeated information.

    \textbf{Somewhat Comprehensible:} The summary makes sense but suffers from grammatical errors, low fluency, and/or repeated information.

    \textbf{Comprehensible:} The summary is understandable. It does not exhibit any grammatical errors, disfluencies, and/or repeated information. \\  \hline 
\multicolumn{1}{c}{} \\ \hline    
    \textbf{Attribution: Is all the information in the summary fully
    attributable to the article?} \\\hline


    \textbf{Yes, it is attributable:} Select this option if it is
    accurate to say, ``The provided news article says…'' or ``According to the news article…'' with the summary following this phrase.
    
    \textbf{No, not fully attributable:} Select this option if only some of the information is supported in the news article, but other parts of the information are missing from the news article or not an accurate representation.
    \\  \hline 
    \multicolumn{1}{c}{} \\ \hline
    \textbf{Informativeness: Is the summary a good summary of the article?} \\ \hline

    \textbf{Bad summary:} The summary does not capture the important information in the article, or the captured information is not accurate with the article. It can also exhibit grammatical issues, low fluency, and/or repeated information.

    \textbf{Good Summary:} The summary captures the important information in the article and presents it accurately and concisely. It does not exhibit any grammatical errors, disfluencies, and/or repeated information. \\
    \hline
    \end{tabular}
    } 
\vspace*{-1.5ex}
    \caption{Instructions used in our  human evaluation.}
    \label{fig:human_eval_instructions} 
    \vspace{-15px}
\end{figure}

\subsection{Human Evaluation}

In addition to automatic evaluation, we conducted a large-scale human
elicitation study assessing different dimensions of the output in all
45 languages. Firstly, we asked participants to read a system summary
and assess its {\em Quality} ({\em Is the summary comprehensible?}),
without looking at the source article, using a~1--3 rating scale where
3~means fully understandable and 1~indicates that the summary has
serious fluency errors.  After the first assessment, participants were
shown the source article and were asked to rate the summary according
to~\emph{Attribution} ({\em Is all the information in the summary
  fully attributable to the article?}) using the attribution
definition\footnote{A fully attributable (or supported)
system-generated summary contains an accurate representation of
information in the source news article. No information in the summary
is unattested when compared against the source news article.} provided
in \citet{ais-paper}; and~\emph{Informativeness} ({\em Is the summary
  a good summary of the article?}). Both assessments used binary
judgements (we report on the percentage of times each system was
rated positively). Figure~\ref{fig:human_eval_instructions} presents our instructions.\footnote{We also
present an example of the interface presented to our participants in
Appendix~\ref{sec:detailed:human:setup}
(Figure~\ref{fig:human_eval_template_full}).}

In order to evaluate summarization output in such a diverse
multilingual setting, we have taken several measures to scale our
study to 45 languages while maintaining high inter-annotator
agreement. We used the same instructions  in English for all languages and
invited bilingual participants (native speakers of the target
language who are also proficient in English) to take part in our
study.\footnote{See Appendix~\ref{sec:detailed:human:setup} for more information on annotators' qualifications and  demographic information.} Each participant had to pass a {\em Screener test}
consisting of 25 questions with an accuracy of 85\% before they could
take part in the study. Finally, we conducted two pilot studies before
the final evaluation to give participants feedback and improve
agreement. 
Our final elicitation study was conducted
using 100 instances per language, each randomly sampled from the test
set.  We collected ratings from three different annotators for each
data point.



\begin{table}[t]
\begin{center}
\small
  \begin{tabular}{@{}l@{}cc} \hline
 & ROUGE & NLI \\ \hline 
  Vanilla & 33.65& 64.40 \\
 Filtered $_{\text{best ROUGE}}$  & \textbf{34.00} & 72.17 \\
 Filtered $_{\text{best NLI}}$  & 32.98 & \textbf{76.49} \\
 Controlled $_{\text{best ROUGE}}$ & 33.00& 72.17 \\ 
 Controlled $_{\text{best NLI}}$ & 33.28 & 71.38 \\ \hline
 mT5-Base \cite{hasan-etal-2021-xl}& 31.85 &  53.41 \\
 Self-ROUGE &33.12 & 67.39  \\
 Random & 33.44 & 69.62  \\ \hline
  \end{tabular}
\end{center}
\vspace*{-1.5ex}
\caption{\label{tab:sum:results} ROUGE-L and NLI scores averaged
  across the 45 languages on the XLSum test set. Highest scores are in bold. Per language results are in Appendix~\ref{sec:detailed:automatic}.}
  \vspace{-15px}
\end{table}

\section{Results}





\paragraph{Filtered Model is Best on NLI-based Evaluation}

Table~\ref{tab:sum:results} presents our results on the test set
averaged across all 45 languages, for our three model variants (Vanilla,
Filtered, and Controlled) and three baselines. For the Filtered and Controlled models we report results for both the best-ROUGE and best-NLI checkpoints, while for the others we only use ROUGE for checkpoint selection as no NLI model is involved in their training. Per language results on the validation and
test sets are in Appendix~\ref{sec:detailed:automatic},
Tables~\ref{tab:dev_results_nli_rouge}
and~\ref{tab:test_results_nli_rouge}.\footnote{Due to the high  computational cost of the experiments, our  results are based on a single run.}

We see that the Filtered model outperforms all other models across
languages achieving
an average score of 76.49  for the Best-NLI checkpoint
(it obtains best NLI scores in 43 out of  45 languages).
This suggests that data filtering is a viable approach for improving
the factual consistency of summarization systems. The next best models
in terms of NLI are the Filtered and Controlled variants (with
Best-ROUGE checkpoints), achieving an average score of~72.17.
The Controlled and Vanilla models perform mostly worse than
the Filtered variant in terms of NLI with either Best-NLI or
Best-ROUGE checkpoints.
Note the significant NLI score gap between the Vanilla
model and the Best-NLI Filtered model (12.19 points on average). This
primarily points to the quality of the unfiltered data, since both
models are based on T5-XXL.
The Best-NLI checkpoint outperforms the Best-ROUGE checkpoint for the
Filtered model (average NLI scores of 76.49 vs 72.17). However, we
observe a degradation of 0.79~NLI points when comparing the Best-NLI
and Best-ROUGE checkpoints for the Controlled model.


\begin{table*}[t!]
\begin{small}
\begin{center}
\resizebox{1.98\columnwidth}{!}{

\begin{tabular}{l|cc|cc|cc|cc|cc|cc} \hline
\multicolumn{1}{l|}{}&\multicolumn{4}{c|}{{Vanilla}} & \multicolumn{4}{c|}{{Filtered}} &\multicolumn{4}{c}{{Controlled}} \\

\multicolumn{1}{l|}{}&\multicolumn{2}{c|}{Best-ROUGE} & \multicolumn{2}{c|}{Best-NLI} &\multicolumn{2}{c|}{Best-ROUGE} & \multicolumn{2}{c|}{Best-NLI} &\multicolumn{2}{c|}{Best-ROUGE} & \multicolumn{2}{c}{Best-NLI}  \\
{Language}  & \multicolumn{1}{l}{ROUGE} & \multicolumn{1}{l|}{NLI} & \multicolumn{1}{l}{ROUGE} &  \multicolumn{1}{l|}{NLI} & \multicolumn{1}{l}{ROUGE} &  \multicolumn{1}{l|}{NLI} & \multicolumn{1}{l}{ROUGE} &  \multicolumn{1}{l|}{NLI} & \multicolumn{1}{l}{ROUGE} &  \multicolumn{1}{l|}{NLI} & \multicolumn{1}{l}{ROUGE} & \multicolumn{1}{l}{NLI} \\

\hline 
English            &  32.51  & 68.31   &  32.93  &  74.23  &  \textbf{33.23}  & 80.32   & 32.4 & \textbf{84.4}      &33.07 &  81.75      & 33.14 & 82.99\\ \hline
\multicolumn{13}{c}{{Training Resources}} \\ \hline
High & 33.21 &	 63.42 & \textbf{33.25} & 67.26 & 32.29 & \textbf{69.24} & 31.23 & 67.79 & 28.53 & 64.79 & 29.37 & 64.80\\
Medium & 33.21 &	 63.22 & 32.37 & 66.37 & \textbf{33.46} & 69.68 & 32.54
& \textbf{74.09} & 32.64 & 69.96 & 32.39 & 67.52 \\
Low & \textbf{34.38} &	 65.89 & 33.69 & 69.00 & 34.03 & 71.98 & 33.03 & \textbf{77.61} & 33.65 & 71.88 & 33.69 & 71.40\\ \hline
\multicolumn{13}{c}{{Language Family}} \\ \hline
Indo-European & 32.62 &	 63.23 & 32.11 & 66.22 & \textbf{33.07} & 69.44 & 32.32
& \textbf{74.48} & 32.02 & 69.61 & 32.20 & 68.18 \\
Romance & 31.61 &	 57.03 & 31.04 & 61.10 & \textbf{31.91} & 66.51 & 31.27 & \textbf{70.21} & 31.00 & 68.83 & 31.48 & 66.09\\
Turkic & 28.90 &	 64.87 & 29.32 & 65.50 & \textbf{29.38} & 70.32 & 28.85 & \textbf{76.35} & 28.25 & 71.19 & 28.99 & 70.18\\
Semitic & 34.78 &	 65.85 & 34.65 & 72.02 & \textbf{34.95} & 74.73 & 33.87
& \textbf{77.62} & 34.13 & 73.00 & 34.60 & 74.36 \\
Afro-Asiatic & \textbf{33.49} &	 62.58 & 32.10 & 67.31 & 32.75 & 71.08 & 32.03
& \textbf{76.07} & 32.71 & 68.31 & 32.53 & 68.89\\
Indo-Iranian & 36.77 &	 67.55 & 37.07 & 68.14 & \textbf{38.04} & 73.39 & 37.10
& \textbf{76.19} & 35.98 & 70.77 & 36.77 & 71.84 \\ \hline
\multicolumn{13}{c}{{XNLI Training Data}} \\ \hline
Available & 33.01 & 62.68 & 33.09 & 67.30 & \textbf{34.03} & 73.53 & 33.06 & \textbf{77.15} & 32.40 & 73.78 & 33.48 & 72.71 \\
Unavailable & 33.90 & 64.97 & 33.21 & 68.03 & \textbf{33.99} & 71.63 &
32.94 & \textbf{76.23} & 33.27 & 71.52 & 33.20 & 70.83\\ \hline
\end{tabular}
} 
\end{center}
\end{small}
\vspace*{-1.5ex}
\caption{ROUGE-L and NLI scores on XLSum test set for best checkpoints
  averaged across language groups. For training resources we consider
  three groups with varying numbers of training examples: High
  ([70K--10K]), Medium ([10K--6K]), and Low (less than 6K). For
  language families, the Indo-European cluster represents Bengali,
  Gujarati, Hindi, russian, Serbian (Cyrillic and Latin), and Sinhala;
  the Romance cluster comprises of French, Protuguese, and Spanish;
  the Turkic cluster contains Azerbaijani, Kyrgyz, Turkish, and Uzpek;
  Semitic languages are Amharic, Arabic, and Tigrinya; the
  Afro-Asiatic cluster groups together Hausa, Oromo, and Somali;
  finally, the Indo-Iranian cluster represents Pashto, Persian, and
  Punjabi; we omit clusters with two members and singletons. We also
  create two subsets depending on whether they
  appear in the XNLI dataset used to train our multilingual NLI model
  (Available, Section~\ref{sec:multnli}) or not
  (Unavailable). Highest ROUGE-L and NLI numbers are in
  bold.}
\label{tab:results_lang_groups:test}
\vspace{-15px}
\end{table*}

\paragraph{Effect of Entailment Signal on ROUGE} 
As shown above, NLI scores improve in all languages when training
uses the signal from the NLI model, either by filtering data or by
using controlled generation. But what is the effect on ROUGE? Looking
at the average ROUGE scores across languages in
Table~\ref{tab:sum:results}, we again see that the best ROUGE is
obtained by the Filtered model, with the Best-ROUGE
checkpoint. Interestingly, this model is trained on much fewer
examples, but obtains better results than the Vanilla and Controlled
variants that use all training examples in XLSum. This
model obtains higher or comparable NLI scores (72.17 and 76.49, for
Best-ROUGE and Best-NLI, respectively) than the other models,
suggesting that it is more accurate with respect to the reference
summaries and more faithful with respect to the input. In general, the Vanilla, Filtered and Controlled models obtain very similar ROUGE scores, ranging between 32.98
and 34.00, while the range of NLI scores is much larger (from~64.30 to
76.49).

\paragraph{Comparison against Baseline Approaches}

Table~\ref{tab:sum:results} also compares to previous work (mT5-Base,
\citealt{hasan-etal-2021-xl}), and the Self-ROUGE and Random selection
baselines. We did not employ any NLI preprocessing in building the
baseline models, neither in filtering or checkpoint selection.
We observe that all model variants (Vanilla, Filtered, and Controlled)
are superior to mT5-Base in terms of ROUGE which is not surprising
given the different model capacities (XXL vs Base). We also see that
\emph{any} filtering improves NLI scores (compare Vanilla against Self-ROUGE
and Random), incurring a slight decrease in terms of ROUGE, while
\emph{targeted} filtering using NLI yields best results. 

\paragraph{ROUGE and NLI across Different Language Groups}

Table~\ref{tab:results_lang_groups:test} shows our results clustered
by (1) the number of training examples per language: High (\mbox{10k--70k} examples), Medium
(6k--10k examples); and Low (less than 6k examples); (2) language
family (Indo-European, Romance, Turkic,
Semitic, Afro-Asiatic, and Indo-Iranian families); and~(3) whether XNLI
training data is available; we cluster languages into two subsets,
those that appear in the XNLI dataset used to train our 
NLI model, and those that do not (see
Table~\ref{tab:training_data}). We report results on English on its
own, as it is the language with the largest number of examples
(370k).

Again, we observe that the Filtered model is in most cases superior,
including English.  Vanilla scores are better on ROUGE for Low resource
and Afro-Asiatic languages, although the difference against other
models is less than 1 ROUGE point. The Controlled model is not better
than Filtered or Vanilla in any configuration, irrespective of how
languages are grouped into clusters.  In conclusion, we find that the
Filtered model dramatically improves faithfulness, while maintaining
ROUGE performance similar to other models. We present examples of
model output in Appendix~\ref{sec:example:output}. 





\paragraph{Human Assessment for Quality, Attribution, and Informativeness}

Table~\ref{tab:final_results} presents our human evaluation results
for Quality, Attribution and Informativeness (it also includes
automatic evaluation results for a side-by-side comparison). We provide per
language analysis in  Appendix~\ref{sec:detailed:human} (See
Tables~\ref{tab:results_human_eval_q1}--\ref{tab:results_human_eval_q3})
and aggregate statistics using the same groups as in
Table~\ref{tab:results_lang_groups:test} (see
Tables~\ref{tab:humaneval_q1_lang_groups}--\ref{tab:humaneval_q3_lang_groups}).

Unsurprisingly, human  reference summaries were more
understandable than Vanilla, Filtered, or Controlled summaries, with
least fluency issues. Differences between the gold standard
summaries and those generated by the Filtered Best-NLI and Controlled
Best-NLI are, however, not statistically significant (using a one-way
ANOVA with post-hoc Tukey HSD tests; \mbox{$p < 0.01$}). Summaries
generated using our Filtered Best-NLI model were most attributed (or
faithful) and informative, with respect to their input
documents. Differences between the Filtered Best-NLI model and all
other comparisons are statistically significant (using a one-way ANOVA
with post-hoc Tukey HSD tests; \mbox{$p < 0.01$}). In conclusion, human
evaluation confirms the Filtered model is best at generating faithful
and informative summaries.

\begin{table*}[t!]
\begin{small}
\begin{center}
\resizebox{1.98\columnwidth}{!}{

\begin{tabular}{l|c|c|c|c|c|c|c} \hline
&\multicolumn{2}{c|}{{Vanilla}} & \multicolumn{2}{c|}{{Filtered}} &\multicolumn{2}{c|}{{Controlled}} &  \multicolumn{1}{c}{Reference}\\

\multicolumn{1}{l|}{{Metric}}&\multicolumn{1}{c|}{{Best-ROUGE}} &
\multicolumn{1}{c|}{{Best-NLI}} &\multicolumn{1}{c|}{{Best-ROUGE}} &
\multicolumn{1}{c|}{{Best-NLI}} &\multicolumn{1}{c|}{{Best-ROUGE}} &
\multicolumn{1}{c|}{{Best-NLI}} & \\

\hline
Quality & 0.85 & 0.84 & 0.85 & 0.86 & 0.84 & 0.86 & \textbf{0.88} \\ 
Attribution & 0.44 & 0.47 & 0.46 & \textbf{0.52} & 0.49 & 0.47 & 0.31  \\
Informativeness & 0.37 & 0.40 & 0.39 & \textbf{0.45} & 0.41 & 0.40 & 0.27 \\\hline
Length Ratio & 0.93 & 0.88 & 0.89 & 0.87 & \textbf{1.00} & 0.81 & 1.00 \\
ROUGE & 33.65 & 33.18 & \textbf{34.00} & 32.98 & 33.02 & 33.28 & --- \\
NLI & 64.31 & 67.82 & 72.18 & \textbf{76.50} & 72.18 & 71.38 & --- \\ \hline

\end{tabular}
} 
\end{center}
\end{small}
\vspace*{-6pt}
\caption{Mean human judgments on XLSum test set averaged across
  languages. We also include ROUGE-L and NLI scores for a side-by-side
  comparison. Length Ratio is the ratio of predicted length to target
  length averaged across all test examples. Best results in each row
  are in \textbf{bold}.}
\label{tab:final_results}
\vspace{-15px}
\end{table*}

\paragraph{Effect on Summary Length}

One may argue that we are improving faithfulness by favoring shorter
summaries. To study this, we also report in
Table~\ref{tab:final_results} the ratio of predicted to target summary
length averaged across all test examples, for different models.

As we can see, best-NLI checkpoints do yield a reduction in predicted
length across different models compared to their Best-ROUGE
checkpoints; the length ratios drop from 0.93 to 0.88 for Vanilla,
from 0.89 to 0.87 for Filtered, and from 1.00 to 0.81 for
Controlled. However, shorter summaries are not necessarily more
faithful; the worst length ratio (0.81) is for the Controlled Best-NLI
model which performs worse on NLI, Attribution, and Informativeness,
compared to the Filtered Best-NLI model with a higher length ratio
(0.87).  The Filtered Best-NLI model only yields a marginal reduction
in summary length compared to the Vanilla Best-Rouge summaries (Length
ratio: 0.87 vs 0.93), but improves on NLI scores (76.50 vs 64.31),
Quality (0.86 vs 0.85), Attribution (0.52 vs 0.44), and
Informativeness (0.45 vs 0.37) assessments.


\section{Metric Learning for Multilingual Summary Evaluation} 

Our large-scale judgment elicitation study (across multiple languages and system outputs) delivered valuable annotations of summary document-quality (31,499 pairs x 3 quality dimensions x 3 raters). We next explore whether it is possible to \emph{learn} metrics for evaluating Quality, Attribution, and Informativeness automatically. Existing metrics (e.g.,~\textsc{Bleurt}; \citealt{sellam-etal-2020-bleurt})  have not targeted summarization specifically, or considered attribution, and multiple languages.  Let $\bm{s} = (s_1, \dots, s_r)$ denote a summary of length~$r$ where each $s_i$
is a token and let
$\bm{d} = (d_1, \dots, d_p)$ be its corresponding input document of length~$p$. Let $\{(d_i, x_i, y_i)\}^{N}_{n=1}$ be a training dataset of
size $N$ where $y_i \in R$ is the human rating that indicates how good~$x_i$ is as a summary of $d_i$ along a specific dimension. Our goal is to learn a function
$\bm{f}: (\bm{d,x}) \rightarrow y$ that predicts the human rating.

We finetuned three models based on mT5-XXL \cite{xue-etal-2021-mt5}, one per dimension (details in Appendix~\ref{sec:detailed:metric:training}). The input  was the concatenation of a document and its summary, and the output the human rating. 10\% of the elicited ratings (across languages) were reserved for testing, while the remainder was used for training and validation. Table~\ref{tab:correlation_analysis} reports correlation coefficients (Pearson's~$r$) between model predictions and (mean) human ratings.
 \textsc{mT5}-Q, \textsc{mT5}-A and \textsc{mT5}-I denote the learned metrics corresponding to Q(uality), A(ttribution), and I(informativeness), respectively. In addition,  we report correlation coefficients for ROUGE and NLI.

\begin{table}[t!]
\begin{small}
\begin{center}
\resizebox{0.98\columnwidth}{!}{
\begin{tabular}{l|c|c|c} \hline
\multicolumn{1}{l|}{{Metric}}&\multicolumn{1}{c|}{{Quality}} & \multicolumn{1}{c|}{{Attribution}} & \multicolumn{1}{c}{{Informativeness}} \\
\hline
ROUGE & 0.12 & 0.15 & 0.12\\
NLI &  0.08 & 0.36 & 0.33\\ 
\textsc{mT5}-Q & \textbf{0.38} & 0.09 &0.14\\
\textsc{mT5}-A & 0.09 &  \textbf{0.57} & 0.52\\
\textsc{mT5}-I & 0.14 &  0.49 & \textbf{0.53} \\
\hline
\end{tabular}
} 
\end{center}
\end{small}
\vspace*{-6pt}
\caption{Correlation of metrics with human summary ratings for the dimensions of Quality (Q), Attribution (A), and Informativeness (I) on the test set. All correlations are statistically significant at $p<0.01$.}
\label{tab:correlation_analysis}
\vspace{-15px}
\end{table}

Overall, we observe that learned metrics correlate best with human ratings (across dimensions). ROUGE correlates weakly with human judgments but cannot distinguish any dimension in particular, whereas NLI scores reliably correlate with attribution. Our results underscore the need for better and more fine-grained evaluation of summary quality, and also corroborate well-known issues \cite{Gehrmann:ea:2022} with widely adopted lexical overlap-based metrics such as ROUGE. 

\section{Conclusion}
In this paper we  leveraged factual consistency evaluation
for improving summarization models in multiple languages.  
Extensive experiments on the XLSum dataset showed large gains when training summarization
models on a subset of the data selected using the NLI signal.  Through a large-scale human evaluation study, we obtained ratings which not only helped us distinguish best performing systems, but were further used to learn metrics for assessing multilingual summaries along the dimensions of Quality, Attribution, and Informativeness. These metrics
could be further used to inspect the quality of summarization datasets.
Our annotators found that summaries are
(on average) only 52\% of the time fully faithful to their documents and 
this number is much worse for some languages (e.g.,~Hausa, Yoruba; see
Table~\ref{tab:results_human_eval_q2} in
Appendix~\ref{sec:detailed:human}). 

An interesting avenue for future
is to directly optimize the summarization models towards the different quality objectives, e.g. via Reinforcement
Learning \cite{narayan-etal-2018-ranking} or Calibrating Sequence
Likelihood \citep{ZhaoSlic}. 


\section*{Limitations}
While our work covers a large number of languages, it is focused on a specific source and style of summaries.
Our experiments focus exclusively on the XLSum dataset \cite{hasan-etal-2021-xl}  which is based on BBC articles where the opening sentence serves as a summary. It would be interesting to explore our methods on additional datasets and text generation tasks, e.g.,~where the summaries are longer, or there are multiple input documents.

\section*{Ethics Statement} An ethical consideration that concerns our work is the problem of misinformation. While we make a step towards improving the factual consistency of text generation systems which in turn should alleviate issues of misinformation, it is important to note that current systems are still far from being perfect in this respect, and thus should be used with caution.

\section*{Acknowledgements}
We thank Ankur Parikh, Sebastian Gehrmann, Dipanjan Das and William Cohen for their feedback on this work. The human rating process was managed by Muqthar Mohammad, Kiranmai Chennuru, Aishwarya Gomatam,  Raghava Ram Pamidigantam and Mahesh Maddinala, without them this work would not have been possible. Thanks for invaluable support from Sheila de Guia and Suneet Dhingra.

\bibliography{anthology,custom} 

\clearpage

\appendix

\begin{table*}[ht!]
\begin{center}
\resizebox{1.98\columnwidth}{!}{
\begin{tabular}{lcccccccccccccccc}
\hline
Model                   & \multicolumn{1}{l}{ar} & \multicolumn{1}{l}{bg} & \multicolumn{1}{l}{de} & \multicolumn{1}{l}{el} & \multicolumn{1}{l}{en} & \multicolumn{1}{l}{es} & \multicolumn{1}{l}{fr} & \multicolumn{1}{l}{hi} & \multicolumn{1}{l}{ru} & \multicolumn{1}{l}{sw} & \multicolumn{1}{l}{th} & \multicolumn{1}{l}{tr} & \multicolumn{1}{l}{ur} & \multicolumn{1}{l}{vi} & \multicolumn{1}{l}{zh} & \multicolumn{1}{l}{Avg.} \\ \hline
\cite{xue-etal-2021-mt5} & 84.5                   & 87.7                   & 87.3                   & 87.3                   & 91.6                   & 87.8                   & 86.9                   & 83.2                   & 85.1                   & 80.3                   & 81.7                   & 83.8                   & 79.8                   & 84.6                   & 83.6                   & 85.0                     \\
Ours                    & 90.0                   & 91.6                   & 91.2                   & 91.4                   & 93.5                   & 91.7                   & 91.4                   & 88.7                   & 90.3                   & 87.2                   & 88.0                   & 89.1                   & 86.3                   & 89.6                   & 89.8                   & 90.0 \\ \hline

\end{tabular}
}
\end{center}
\caption{Accuracy results on XNLI test set.}
\label{tab:xnli_full}
\end{table*}

\begin{table*}[ht!]
\begin{center}
\resizebox{1.98\columnwidth}{!}{
\begin{tabular}{lcccccccccccc}
\hline
Model   & \multicolumn{1}{l}{FRANK} & \multicolumn{1}{l}{MNBM} & \multicolumn{1}{l}{QAGS-C} & \multicolumn{1}{l}{QAGS-X} & \multicolumn{1}{l}{BEGIN} & \multicolumn{1}{l}{$Q^2$} & \multicolumn{1}{l}{DialFact} & \multicolumn{1}{l}{PAWS} & \multicolumn{1}{l}{FEVER} & \multicolumn{1}{l}{VitaminC} & \multicolumn{1}{l}{SummEval} & \multicolumn{1}{l}{Avg.} \\ \hline
~\cite{honovich-etal-2022-true} & 89.4                      & 77.9                     & 82.1                       & 83.8                       & 82.6                      & 72.7                                     & 77.7                         & 86.4                     & 93.2                      & 88.3                         & 80.5                         & 83.1                     \\
Ours    & 88.1                      & 77.2                     & 82.8                       & 84.5                       & 80.5                      & 72.8                                     & 80.1                         & 82.6                     & 93.7                      & 88.2                         & 75.6                         & 82.4  \\ \hline   

\end{tabular}
}
\end{center}
\caption{ROC AUC results on TRUE.}
\label{tab:true_full}
\end{table*}

\section{Intrinsic NLI Model Evaluation}
\label{sec:appendix_nli}

In this section we present more detailed evaluation results for our
our multilingual NLI model.  Table~\ref{tab:xnli_full} shows accuracy
the on XNLI test set for 15~languages; we compare our results to the
model of \citet{xue-etal-2021-mt5}; it is also based on mT5-XXL
finetuned on MNLI.  The table shows our model achieves higher
accuracy for all languages, however, results are not fully comparable
as we only consider binary labels (entailment/non-entailment) in our
setting in comparison to three classes
(entailment/neutral/contradiction) for \citet{xue-etal-2021-mt5}.

We additionally evaluate our model on the TRUE factual consistency
benchmark~\cite{honovich-etal-2022-true}. TRUE consists of 11 diverse
datasets (including the output of grounded text generation systems),
annotated with binary factual consistency labels. Although TRUE only
includes English examples, we use it for our evaluation due to its
relevance to factual consistency in
summarization. Table~\ref{tab:true_full} shows the area under the ROC
curve (ROC AUC) results for all dataset in TRUE where we compare our
multilingual model to T5-11B trained on
ANLI~\cite{nie-etal-2020-adversarial} reported by
\citet{honovich-etal-2022-true}. Results show that our multilingual model
performs on par with theirs,  while finetuning our model on
non-English data  causes a slight decrease in performance.

\section{Technical Modeling Details}
\label{sec:appendix_technical}
We used the \emph{t5x} \cite{roberts2022scaling} framework for all training and inference tasks. We ran all experiments on TPU accelerators.

\section{Detailed Automatic Evaluation Results}
\label{sec:detailed:automatic}

Table~\ref{tab:dev_results_nli_rouge} (development set) and
Table~\ref{tab:test_results_nli_rouge} (test set) report ROUGE-L and
NLI scores on XLSum broken for individual
languages. Table~\ref{tab:test_results_additional baselines} compares
our Filtered model against previous work (mT5-Base,
\citealt{hasan-etal-2021-xl}) and the Self-ROUGE and Random selection
baselines. Results are presented for individual languages and on
average on the XLSum \emph{test set}.

Table~\ref{tab:language-family} shows details of how we grouped
languages into different clusters (i.e., family and the availability
of NLI training data). Table~\ref{tab:results_families} shows our
results on the XLSum \emph{development set} clustered by (1) the number of
training examples per language; we group languages into three clusters
-- High (\mbox{10k--70k} examples), Medium (9k--16k examples); and Low (less
than 6k examples); (2) language family (we group languages into
Indo-European, Romance, Turkic, Semitic, Afro-Asiatic, and
Indo-Iranian families); and~(3) whether XNLI training data is
available; we cluster languages into two subsets, those that appear in
the XNLI dataset used to train our multilingual NLI model, and those
that do not (see Table~\ref{tab:training_data}). We report results on
English on its own, as it is the language with the largest number of
examples (370k).

\section{Human Evaluation Setup and Annotator Qualifications}
\label{sec:detailed:human:setup}

\begin{figure*}[t!]
    \centering
    \footnotesize
    \begin{tabular}{p{15.5cm}}
    \frame{\includegraphics[scale=0.25]{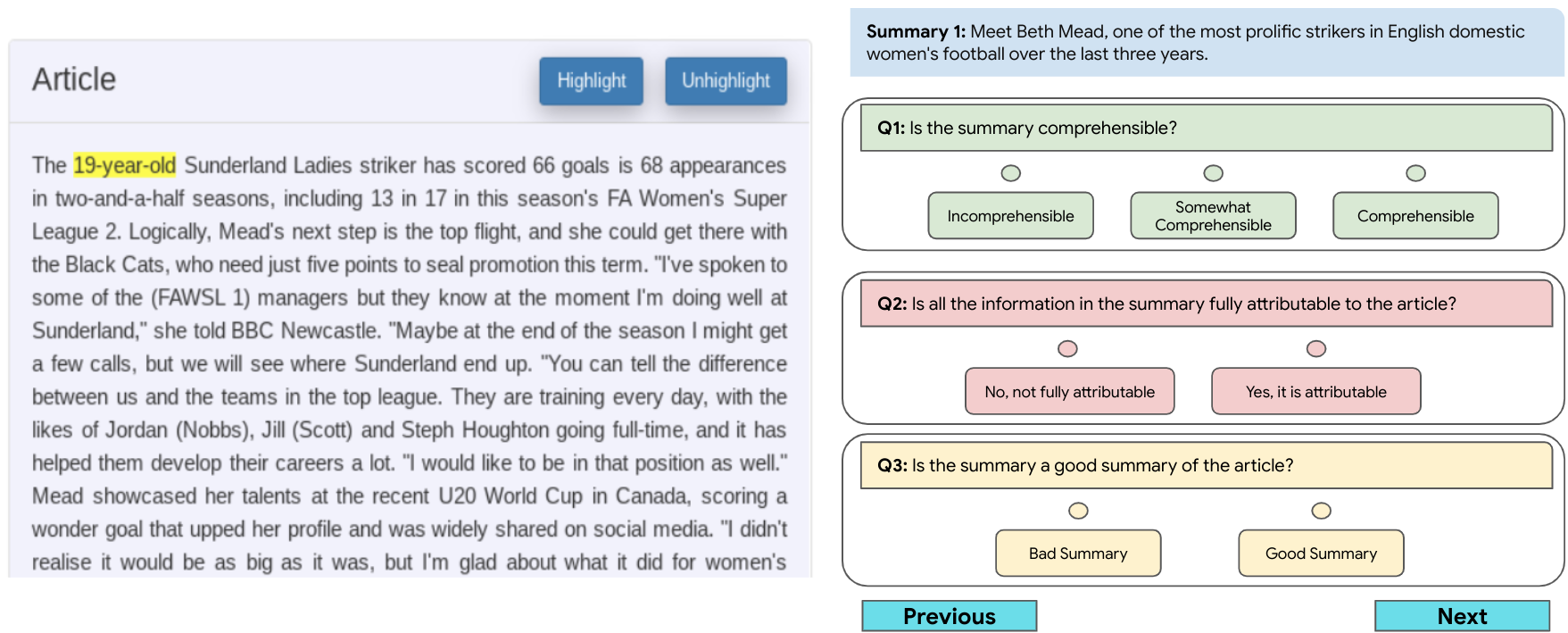}} \\   \vspace{0.1cm}
    \end{tabular}
    
    \begin{tabular}{lp{11cm}} \hline
    \multicolumn{2}{c}{Q1: Is the summary comprehensible?} \\
 \hline   
    \textbf{Incomprehensible:} & The summary is difficult to understand. It can have serious grammatical errors, low fluency, and/or repeated information.\\

    \textbf{Somewhat Comprehensible:} &The summary generally makes sense but suffers from grammatical errors, low fluency, and/or repeated information.\\

    \textbf{Comprehensible:} & The summary is understandable. It does not exhibit any grammatical errors, disfluencies, and/or repeated information. \\ \hline 
\multicolumn{2}{c}{} \\

\hline
    \multicolumn{2}{c}{Q2: Is all the information in the summary fully
      attributable to the article?} \\ \hline

    \textbf{Yes, it is attributable:}  &Select this option if it is
    accurate to say, “The provided news article says…” or “According
    to the news article…” with the summary following this phrase. \\
    
    \textbf{No, not fully attributable:} & Select this option if only
    some of the information is supported in the news article, but
    other parts of the information are missing from the news article
    or not an accurate representation.    \\
    \hline
\multicolumn{2}{c}{} \\ \hline
    \multicolumn{2}{c}{Q3: Is the summary a good summary of the article?}\\

    \textbf{Bad summary:} &The summary does not capture the important
    information in the article, or the captured information is not
    accurate with the article. It can also exhibit grammatical issues,
    low fluency, and/or repeated information. \\

    \textbf{Good Summary:} &The summary captures the important
    information in the article and presents it accurately and
    concisely. It does not exhibit any grammatical errors,
    disfluencies, and/or repeated information.\\ \hline

    \end{tabular}
    \caption{A snapshot of the interface and  instructions  were used
      in our human evaluation studies.}
    \label{fig:human_eval_template_full}
\end{figure*}

Figure~\ref{fig:human_eval_template_full} presents a snapshot of the
interface seen by our participants together with the instructions used
in our human evaluation studies.

To recruit our participants, we screened their language skills to determine whether they're native speakers, their education level and country of residence as well as origin. For some languages we could not recruit native speakers in the country of birth for various restrictions and sourcing difficulties, we hired native speakers in other countries. In addition, we created a screener test to determine the raters' suitability for the task. In total, we recruited 388 raters across all 45 locales. 2.58\% of them hold a Doctorate, 31.96\% holds a master degree, 57.73\% of them hold a bachelor degree, 7.73\% hold High school degree or equivalent. Table~\ref{tab:annotator-demo} presents the demographics of our participants. All our annotators are paid adequately by our suppliers adhering to the supplier code of conduct. 

\section{Detailed Human Evaluation Results}
\label{sec:detailed:human}

Table~\ref{tab:humaneval_q1_lang_groups} presents human
evaluation results for Summary Quality for individual languages on the
XLSum test set. Table~\ref{tab:results_human_eval_q2} shows mean
judgments for Attribution, again per language, and finally
Table~\ref{tab:results_human_eval_q2} summarizes our results for
Informativeness. We also group human judgments according to number of
training examples, language family, and whether XNLI training data is
available. Tables~\ref{tab:humaneval_q1_lang_groups}--\ref{tab:humaneval_q3_lang_groups} show these different types of clustering for
the judgments pertaining to Summary Quality, Attribution, and
Informativeness.

\section{Metric Training Details}
\label{sec:detailed:metric:training}

The metrics were trained by finetuning mT5-XXL to predict a binarized version of the human judgments (a summary receives a score of 1 if the mean human rating $>0.5$). Each metric is trained for 20,000 steps with batch size $=32$ and a learning rate of $0.0001$. Checkpoints were selected by their accuracy on a validation set.

\section{Example Output}
\label{sec:example:output}

We showcase summaries generated by our models in
Tables~\ref{tab:output:fr1}, ~\ref{tab:output:fr2}, and ~\ref{tab:output:es}. The article in
Table~\ref{tab:output:fr1} discusses a cholera outbreak in Algeria,
with two deaths, 46 confirmed cases and 88 suspected cases. The
reference summary  in addition mentions that there have been 139
hospitalizations since August 2018, however, the number of
hospitalizations is not given in the input document. Tha Vanilla
summary manages to hallucinate two facts: the deaths have not been
\emph{several}, they are only two, and the number of suspected cases
is 88, not 100. The Controlled summary is factual although perhaps 
sparse with the details, it only mentions the cholera deaths but not
the cases. The Filtered summary on the other hand correctly mentions
the number of  confirmed and suspected cases but does not mention the
deaths. 

The article in Table~\ref{tab:output:fr2}  talks about trials of an
Ebola vaccine in Oxford. The trial involves 72 volunteers, and
preliminary tests on monkeys have shown that the vaccine confers
immunity against Ebola. Similar small-scale trials are underway in the
United states and three African countries spared from the
epidemic. The Reference summary is factually correct, the Vanilla
summary gives the false impression that the Ebola trial is
large-scale;  by omitting the adjective ``large-scale'', the
Controlled summary is factual, and likewise the Filtered summary does
not include any hallucinations. 

The article in Table~\ref{tab:output:es} talks about Greenpeace activists
arrested in Russia on piracy charges for protesting against and oil rig in the arctic sea.
Among the 30 arrested, two are Argentinian, one Brazilian.
Five of them were accused of climbing the oil rig and their detention was extended by two months.
The Reference summary is factually correct, the Vanilla
summary has slight fluency issues, it hallucinates the location of the oil rig to be in the Black Sea,
and misrepresents the protest to be about the oil rig closure;
the Controlled summary is factual but focuses on the oil rig climbers, and likewise the Filtered summary does
not contain any hallucinations but focuses on the Argentine activists only.

\begin{table*}[ht!]
\begin{small}
\begin{center}
\resizebox{1.98\columnwidth}{!}{
\rowcolors{1}{}{lightgray}
\begin{tabular}{l|c|l|c|l|c|l|c|l|c|l|c|l}
\rowcolor{white}
\multicolumn{1}{l|}{}&\multicolumn{4}{c|}{\textbf{Vanilla}} & \multicolumn{4}{c|}{\textbf{Filtered}} &\multicolumn{4}{c}{\textbf{Controlled}} \\

\rowcolor{white}
\multicolumn{1}{l|}{}&\multicolumn{2}{c|}{\textbf{Best-ROUGE}} & \multicolumn{2}{c|}{\textbf{Best-NLI}} &\multicolumn{2}{c|}{\textbf{Best-ROUGE}} & \multicolumn{2}{c|}{\textbf{Best-NLI}} &\multicolumn{2}{c|}{\textbf{Best-ROUGE}} & \multicolumn{2}{c}{\textbf{Best-NLI}}  \\

\textbf{Language}  & \multicolumn{1}{l|}{\textbf{ROUGE}} & \multicolumn{1}{l|}{\textbf{NLI}} & \multicolumn{1}{l|}{\textbf{ROUGE}} &  \multicolumn{1}{l|}{\textbf{NLI}} & \multicolumn{1}{l|}{\textbf{ROUGE}} &  \multicolumn{1}{l|}{\textbf{NLI}} & \multicolumn{1}{l|}{\textbf{ROUGE}} &  \multicolumn{1}{l|}{\textbf{NLI}} & \multicolumn{1}{l|}{\textbf{ROUGE}} &  \multicolumn{1}{l|}{\textbf{NLI}} & \multicolumn{1}{l|}{\textbf{ROUGE}} & \multicolumn{1}{l}{\textbf{NLI}} \\

\hline
Amharic            &   33.54   &  60.4   &   33.02  & 65.43 &   33.17  & 66.47           & 32.72  & \textbf{73.38} &   \textbf{33.57}    & 62.12          &  32.94    & 66.09 \\
Arabic             &   29.94   &  38.2   &   31.94  & 53.01 &   \textbf{32.49}  & 61.29           & 32.11  & \textbf{65.29} &   30.58    & 52.63          &  31.91    & 58.97 \\
Azerbaijani        &   \textbf{29.05}   &  53.14  &   28.76  & 51.56 &   28.79  & 61.14           & 28.39  & \textbf{67.39} &   27.84    & 53.96          &  28.25    & 55.29 \\
Bengali            &   \textbf{33.96}   &  62.85  &   33.22  & 61.61 &   33.36  & 70.6            & 33.61  & \textbf{77.53} &   33.68    & 59.77          &  32.88    & 67.1  \\
Burmese            &   \textbf{40.88}   &  57.62  &   39.6  & 60.48 &   39.19  & 70.19           & 38.91  & \textbf{76.96} &   39.41    & 62.03          &  39.4    & 66.49 \\
Chinese (simp.)    &   32.44   &  47.32  &   32.76  & 58.2  &   \textbf{34.93}  & 71.09           & 34.31  & \textbf{72.99} &   32.61    & 62.88          &  33.64    & 67.46 \\
Chinese (trad.)    &   33.57   &  52.99  &   34.11  & 58.7  &   \textbf{36.3}  & \textbf{72.42}  & 34.98  & 71.09          &    33.28   & 62.3           &   34.35   & 66.06 \\
English            &   31.68   &  55.52  &   32.26  & 70.7  &   \textbf{32.37}  & 72.83           & 32.3   & \textbf{80.48} &   32.61    & 69.61          &  32.44    & 74.58 \\
French             &   \textbf{33.62}   &  51.19  &   33.04  & 51.02 &   33.13  & 61.24           & 32.94  & \textbf{64.92} &   32.6    & 60.07          &  32.82    & 58.26 \\
Gujarati           &   \textbf{32.83}   &  50.1   &   31.93  & 48.79 &   32.04  & 55.54           & 32.25  & \textbf{61.14} &   32.45    & 53.92          &  31.59    & 52.64 \\
Hausa              &   \textbf{37.34}   &  48.85  &   35.64  & 51.52 &   36.74  & 56.83           & 35.96  & \textbf{62.81} &   36.93    & 54.71          &  35.89    & 57.34 \\
Hindi              &   35.53   &  56.8   &   35.67  & 62.27 &   \textbf{36.39}  & \textbf{68.92}  & 36.1   & 68.31          &   35.62    & 62.78          &  36.15    & 62.02 \\
Igbo               &   35.09   &  44.85  &   35.23  & 48.13 &   35.36  & 57.84           & 35.14  & \textbf{59.15} &   34.98    & 48.25          &  \textbf{35.6}    & 52.89 \\
Indonesian         &   33.06   &  57.62  &   33.32  & 60.29 &   \textbf{34.28}  & 69.7            & 33.32  & \textbf{70.54} &   33.3    & 63.68          &  33.38    & 68.13 \\
Japanese           &   40.3   &  68.48  &   39.82  & 72.57 &   \textbf{40.9}  & 76.52           & 39.7   & 77.08          &    39.92   & \textbf{81.95} &   39.54   & 81.07 \\
Kirundi            &   32.37   &  47.9   &   31.32  & 46.54 &   32.44  & 52.65           & 31.2   & \textbf{60.33} &   \textbf{33.15}    & 49.56          &  31.74    & 51.04 \\
Korean             &   39.83   &  66.88  &   37.21  & 67.36 &   38.96  & 70.23           & 37.75  & \textbf{79.42} &   \textbf{39.89}    & 70.31          &  36.29    & 74.05 \\
Kyrgyz             &   25.8   &  53.27  &   \textbf{26.37}  & 50.89 &   24.78  & 56.52           & 25.29  & \textbf{68.1}  &   25.29    & 58.87          &  25.62    & 60.43 \\
Marathi            &   28.62   &  47.15  &   27.2  & 52.72 &   30.09  & \textbf{63.59}  & \textbf{30.84}  & 63.35          &   29.32    & 49.29          &  26.81    & 52.68 \\
Nepali             &   37.96   &  67.53  &   37.48  & 67.11 &   \textbf{38.14}  & 72.3            & 37.06  & \textbf{74.48} &   37.4    & 67.72          &  36.32    & 71.51 \\
Oromo              &   \textbf{29.63}   &  59.69  &   27.37  & 63.12 &   28.83  & 68.41           & 28.07  & \textbf{75.19} &   28.61    & 59.08          &  28.19    & 61.31 \\
Pashto             &   38.96   &  58.37  &   38.57  & 57.27 &   \textbf{39.55}  & 70.66           & 38.73  & \textbf{73.06} &   38.19    & 60.74          &  38.17    & 65.73 \\
Persian            &   34.91   &  60.31  &   36.35  & 64.62 &   \textbf{37.37}  & 71.47           & 36.83  & \textbf{72.66} &   35.55    & 63.05          &  35.99    & 68.45 \\
Pidgin             &   34.27   &  51.04  &   33.78  & 60.66 &   \textbf{34.5}  & 67.76           & 33.27  & \textbf{67.96} &    34.43   & 57.5           &   34.1   & 61.76 \\
Portuguese         &   32.33   &  39.16  &   32.21  & 48.18 &   \textbf{33.43}  & 57.06           & 32.66  & \textbf{57.37} &   32.4    & 48.76          &  32.67    & 57.22 \\
Punjabi            &   \textbf{34.88}   &  48.49  &   33.82  & 48.97 &   34.41  & 50.05           & 34.53  & \textbf{55.12} &   34.46    & 45.76          &  33.87    & 50.55 \\
Russian            &   27.33   &  47.09  &   28.02  & 52.99 &   \textbf{28.06}  & 56.07           & 27.94  & \textbf{61.11} &   27.89    & 59.66          &  27.95    & 61.59 \\
Scottish Gaelic    &   \textbf{33.7}   &  50.55  &   33.36  & 61.03 &   32.38  & 56.82           & 32.63  & \textbf{72.31} &   33.24    & 63.18          &  32.71    & 62.77 \\
Serbian (Cyrillic) &   \textbf{27.76}   &  47.21  &   27.14  & 48.95 &   27.69  & 55.19           & 26.78  & \textbf{57.38} &   27.34    & 50.49          &  26.55    & 51.76 \\
Serbian (Latin)    &   27.15   &  38.59  &   26.84  & 47.07 &   \textbf{27.95}  & 50.49           & 26.25  & \textbf{55.74} &   27.14    & 44.18          &  25.19    & 41.25 \\
Sinhala            &   \textbf{36.16}   &  57.9   &   34.36  & 60.04 &   34.67  & 61.27           & 34.24  & \textbf{76.16} &   35.3    & 66.45          &  34.61    & 69.81 \\
Somali             &   30.93   &  51.75  &   30.46  & 60.48 &   \textbf{31.57}  & 62.42           & 30.52  & \textbf{69.28} &   30.9    & 57.59          &  30.76    & 57.92 \\
Spanish            &   26.63   &  38.53  &   26.25  & 42.13 &   \textbf{26.88}  & 48.84           & 26.48  & \textbf{61.87} &   26.67    & 46.36          &  26.65    & 44.34 \\
Swahili            &   \textbf{34.95}   &  56.8   &   34.83  & 60.1  &   34.91  & 66.87           & 33.66  & \textbf{71.64} &   34.74    & 62.74          &  33.61    & 61.51 \\
Tamil              &   30.82   &  61.6   &   29.5  & 70.35 &   \textbf{31.98}  & 78.94           & 30.46  & \textbf{81.49} &   30.22    & 75.1           &  29.81    & 75.82 \\
Telugu             &   27.9   &  54.06  &   27.94  & 56.94 &   28.01  & 61.29           & 28.29  & \textbf{69.97} &   \textbf{28.47}    & 54.84          &  27.53    & 58.79 \\
Thai               &   29.46   &  54.3   &   \textbf{30.22}  & 61.89 &   29.2  & 68.32           & 29.19  & \textbf{72.93} &    29.08   & 60.84          &   29.78   & 65.0  \\
Tigrinya           &   \textbf{35.25}   &  57.08  &   33.9  & 63.66 &   35.11  & 67.61           & 32.9   & \textbf{73.36} &   33.41    & 52.28          &  34.69    & 60.57 \\
Turkish            &   29.94   &  50.76  &   30.75  & 54.47 &   \textbf{32.18}  & 66.95           & 31.39  & \textbf{68.55} &   30.64    & 57.64          &  31.09    & 60.72 \\
Ukrainian          &   28.16   &  44.18  &   28.6  & 52.46 &   \textbf{28.75}  & 61.42           & 27.98  & \textbf{63.55} &   28.43    & 56.66          &  28.49    & 60.08 \\
Urdu               &   36.03   &  52.54  &   35.79  & 52.69 &   \textbf{37.34}  & 66.87           & 36.75  & \textbf{70.13} &   36.02    & 59.9           &  36.15    & 62.75 \\
Uzbek              &   28.1   &  60.46  &   27.94  & 59.44 &   \textbf{28.12}  & 61.3            & 27.75  & \textbf{70.78} &   27.7    & 59.99          &  27.46    & 59.42 \\
Vietnames          &   34.71   &  53.8   &   34.65  & 59.88 &   \textbf{35.48}  & 67.14           & 34.95  & \textbf{68.21} &   34.38    & 59.73          &  35.1    & 64.89 \\
Welsh              &   \textbf{34.56}   &  55.44  &   32.7  & 52.02 &   33.64  & \textbf{66.7}   & 32.21  & 62.06          &   33.5    & 57.04          &  32.46    & 60.19 \\
Yoruba             &   \textbf{37.25}   &  54.79  &   35.76  & 54.26 &   36.66  & 58.31           & 36.04  & \textbf{66.82} &   36.82    & 50.83          &  36.03    & 55.87 \\
\hline  
Average            &   32.87   &  53.18  &  32.47   & 57.16 &  \textbf{33.17}   & 63.91  & 32.56  & \textbf{68.65}          & 32.67      & 58.59          &   32.38     & 61.43 \\
\end{tabular}
} 
\end{center}
\end{small}
\caption{ROUGE-L and NLI scores per language on the XLSum development
  set for the Best-ROUGE and Best-NLI checkpoints (chosen by averaging across all languages). Highest scores in each row are in \textbf{bold}.}
\label{tab:dev_results_nli_rouge}
\end{table*}

\begin{table*}[ht!]
\begin{small}
\begin{center}
\resizebox{1.98\columnwidth}{!}{
\rowcolors{1}{}{lightgray}
\begin{tabular}{l|c|c|c|c|c|c|c|c|c|c|c|c}

\rowcolor{white}
\multicolumn{1}{l|}{}&\multicolumn{4}{c|}{\textbf{Vanilla}} & \multicolumn{4}{c|}{\textbf{Filtered}} &\multicolumn{4}{c}{\textbf{Controlled}} \\

\rowcolor{white}
\multicolumn{1}{l|}{}&\multicolumn{2}{c|}{\textbf{Best-ROUGE}} & \multicolumn{2}{c|}{\textbf{Best-NLI}} &\multicolumn{2}{c|}{\textbf{Best-ROUGE}} & \multicolumn{2}{c|}{\textbf{Best-NLI}} &\multicolumn{2}{c|}{\textbf{Best-ROUGE}} & \multicolumn{2}{c}{\textbf{Best-NLI}}  \\

\textbf{Language}  & \multicolumn{1}{l|}{\textbf{ROUGE}} & \multicolumn{1}{l|}{\textbf{NLI}} & \multicolumn{1}{l|}{\textbf{ROUGE}} &  \multicolumn{1}{l|}{\textbf{NLI}} & \multicolumn{1}{l|}{\textbf{ROUGE}} &  \multicolumn{1}{l|}{\textbf{NLI}} & \multicolumn{1}{l|}{\textbf{ROUGE}} &  \multicolumn{1}{l|}{\textbf{NLI}} & \multicolumn{1}{l|}{\textbf{ROUGE}} &  \multicolumn{1}{l|}{\textbf{NLI}} & \multicolumn{1}{l|}{\textbf{ROUGE}} & \multicolumn{1}{l}{\textbf{NLI}} \\

\hline
Amharic            &  \textbf{35.60}  & 72.56 &  34.91           &  78.71  &  35.15          & 74.65   & 34.55 & \textbf{79.74}    &35.1 &  75.52     & 35.22 & 79.12\\ 
Arabic             &  32.00           & 55.15 &  33.31           &  64.02  &  33.33          & 72.83   & 33.05 & \textbf{74.88}    &31.67 &  71.11  & \textbf{33.34} & 71.68\\ 
Azerbaijani        &  \textbf{29.95} & 62.15 &  29.07           &  62.2   &  29.28          & 68.03   & 29.26 & \textbf{73.41}    &27.99 &  66.75     & 29.0 & 67.03\\
Bengali            &  \textbf{34.19} & 71.12 &  33.71           &  74.49  &  34.17          & 76.53   & 33.38 & \textbf{81.14}    &34.14 &  73.64     & 33.81 & 73.48\\
Burmese            &  \textbf{41.40}  & 68.54 &  39.83           &  69.56  &  40.58          & 73.68   & 39.08 & \textbf{76.26}    &40.55 &  70.74     & 40.78 & 74.32\\
Chinese (simp.)    &  33.54          & 62.43 &  33.08           &  69.07  &  \textbf{35.66} & 76.95   & 34.15 & \textbf{81.3}     &32.48 &  78.83     & 34.44 & 77.72\\
Chinese (trad.)    &  34.39          & 63.10 &  34.37           &  69.02  &  \textbf{36.49} & 78.15   & 34.43 & \textbf{79.48}    &32.86 &  77.50     & 34.95 & 77.95\\
English            &  32.51          & 68.31 &  32.93           &  74.23  &  \textbf{33.23} & 80.32   & 32.4 & \textbf{84.4}      &33.07 &  81.75      & 33.14 & 82.99\\ 
French             &  34.12          & 64.88 &  33.87           &  67.61  &  \textbf{34.49} & 72.94   & 33.98 & \textbf{76.4}     &33.18 &  77.11     & 34.35 & 72.59\\
Gujarati           &  \textbf{33.21} & 62.52 &  32.19           &  62.12  &  32.59          & 63.54   & 32.46 & \textbf{71.64}    &32.23 &  62.38     & 32.74 & 62.85\\ 
Hausa              &  \textbf{38.02} & 60.59 &  36.51           &  63.07  &  37.3           & 67.47   & 36.75 & \textbf{72.18}     &37.25 &  66.37      & 36.97 & 64.58\\
Hindi              &  36.88          & 66.74 &  36.92           &  69.77  &  \textbf{37.45} & 74.20   & 36.97 & \textbf{77.26}    &36.17 &  73.31     & 37.26 & 74.70\\
Igbo               &  34.17          & 56.01 &  \textbf{35.45}  &  60.22  &  35.44          & 64.36   & 34.19 & \textbf{72.00}       &35.3 &  66.77     & 35.32 & 70.01\\ 
Indonesian         &  34.47          & 67.63 &  34.05           &  72.5   &  \textbf{35.26} & 79.06   & 33.76 & \textbf{82.31}    &34.05 &  78.27     & 34.66 & 78.65\\
Japanese           &  41.19          & 76.81 &  40.86           &  79.35  &  \textbf{41.30}  & 78.60   & 39.52 & 81.29     &41.0 &  \textbf{87.84}      & 40.65 & 85.60\\
Kirundi            &  \textbf{33.70}  & 65.50 &  32.67           &  66.31  &  33.63          & 70.66   & 31.74 & \textbf{75.37}    &33.05 &  72.57     & 32.73 & 67.72\\ 
Korean             &  \textbf{40.36} & 72.05 &  39.28           &  74.00     &  39.83          & 78.51   & 38.24 & \textbf{82.18} &39.9 &  80.02  & 39.28 & 78.12\\ 
Kyrgyz             &  26.48          & 67.57 &  \textbf{27.02}  &  65.58  &  26.54          & 67.99   & 25.41 & \textbf{78.02}    &26.82 &  73.01     & 26.70 & 73.39\\
Marathi            &  30.11          & 61.84 &  28.08           &  64.23  &  \textbf{31.31} & 67.86           & 30.95   & \textbf{70.41}    &28.85 &  66.95     & 27.75 & 63.77\\
Nepali             &  \textbf{39.25} & 76.43 &  38.8            &  78.46  &  38.67          & 80.57   & 37.48 & \textbf{85.65}    &38.42 &  80.86     & 37.40 & 77.18\\
Oromo              &  \textbf{30.67} & 68.92 &  28.76           &  72.55  &  29.61          & 76.35   & 28.76 & \textbf{79.62}    &29.53 &  71.34     & 29.62 & 73.94\\
Pashto             &  39.07          & 70.01 &  39.04           &  70.9   &  \textbf{39.88} & 76.34    & 38.83 & \textbf{80.21}   &38.45 &  72.46    & 38.85 & 73.72\\
Persian            &  35.11          & 71.68 &  36.47           &  71.57  &  \textbf{37.86} & 79.22   & 36.73 & \textbf{80.7}     &34.16 &  78.50     & 35.82 & 80.23\\
Pidgin             &  34.33          & 62.07 &  32.95           &  67.91  &  \textbf{34.41} & 72.03   & 33.07 & \textbf{76.32}    &34.07 &  70.35     & 33.82 & 69.85\\ 
Portuguese         &  33.3           & 52.89 &  32.88           &  57.12  &  \textbf{33.9}  & 64.24   & 33.14 & 64.91   &32.76 &  68.01  & 33.16 & \textbf{68.12}\\
Punjabi            &  36.12          & 60.96 &  35.71           &  61.95  &  \textbf{36.39} & 64.61   & 35.73 & \textbf{67.67}    &35.33 &  61.34     & 35.64 & 61.57\\
Russian            &  28.48          & 58.42 &  28.38           &  62.88  &  \textbf{28.71} & 69.20   & 28.26 & \textbf{72.51}    &28.09 &  71.88     & 28.58 & 70.54\\
Scottish Gaelic    &  33.41          & 57.38 &  33.01           &  66.48  &  32.95          & 68.14   & 32.71 & \textbf{76.06}    &\textbf{33.47} &  71.90     & 32.57 & 66.07\\
Serbian (Cyrillic) &  28.31          & 56.71 &  27.32           &  62.36  &  \textbf{29.20}  & 62.97   & 27.93 & \textbf{69.32}     &28.14 &  66.63      & 27.68 & 62.71\\
Serbian (Latin)    &  26.67          & 51.23 &  25.49           &  54.11  &  \textbf{28.21} & 59.93   & 26.84 & \textbf{65.08}    &25.74 &  62.00     & 24.09 & 51.26\\
Sinhala            &  \textbf{36.33} & 71.25 &  35.72           &  74.55  &  36.04          & 74.76   & 35.67 & \textbf{81.58}    &35.53 &  73.61     & 36.03 & 74.28\\
Somali             &  \textbf{31.77} & 58.23 &  31.04           &  66.32  &  31.35          & 69.42   & 30.59 & \textbf{76.42}    &31.36 &  67.22     & 30.99 & 68.14\\ 
Spanish            &  27.40          & 53.33 &  26.38           &  58.58  &  \textbf{27.33} & 62.34   & 26.7 & \textbf{69.32}     &27.06 &  61.36      & 26.92 & 57.55\\
Swahili            &  35.64          & 64.57 &  35.93           &  69.28  &  \textbf{36.46} & 73.25   & 35.23 & \textbf{78.58}    &35.59 &  73.35     & 36.05 & 68.49\\
Tamil              &  31.94          & 70.37 &  30.31           &  77.9   &  \textbf{33.26} & 81.91    & 30.99 & \textbf{83.11}    &31.49 &  82.14     & 30.93 & 81.76\\
Telugu             &  \textbf{29.35} & 60.61 &  28.61           &  67.74  &  29.11          & 67.83   & 28.88 & \textbf{75.10}     &29.11 &  66.30     & 28.4 & 66.54\\ 
Thai               &  30.59          & 61.83 &  29.92           &  68.77  &  \textbf{30.66} & 73.27   & 29.63 & \textbf{78.95}    &29.1 &  74.01     & 30.42 & 71.33\\
Tigrinya           &  \textbf{36.73} & 69.83 &  35.73           &  73.32  &  36.37          & 76.71   & 34.01 & \textbf{78.24}    &35.63 &  72.37     & 35.24 & 72.27\\
Turkish            &  30.80          & 62.36 &  32.19           &  64.58  &  \textbf{33.65} & 74.66   & 32.45 & \textbf{75.65}    &30.81 &  74.60     & 32.5 & 72.57\\
Ukrainian          &  28.34          & 58.83 &  28.61           &  63.15  &  \textbf{28.99} & 70.87   & 27.66 & \textbf{73.64}    &28.01 &  72.41     & 28.42 & 72.61\\
Urdu               &  36.93          & 67.84 &  37.12           &  69.5   &  \textbf{38.18} & 74.81    & 37.05 & \textbf{77.31}    &36.14 &  73.41     & 37.45 & 75.64\\
Uzbek              &  28.37          & 67.41 &  \textbf{29.02}  &  69.66  &  28.06          & 70.62   & 28.27 & \textbf{78.32}    &27.39 &  70.40     & 27.76 & 67.74\\
Vietnames          &  35.91          & 65.94 &  35.77           &  67.54  &  \textbf{36.69} & 73.01   & 35.46 & \textbf{76.94}    &34.99 &  70.91     & 35.86 & 71.52\\
Welsh              &  \textbf{35.31} & 65.27 &  33.75           &  65.32  &  33.89          & 76.06   & 32.47 & \textbf{76.11}    &34.33 &  71.58     & 33.66 & 71.47\\
Yoruba             &  \textbf{37.65} & 63.91 &  36.01           &  63.22  &  37.01          & 68.48   & 35.15 & \textbf{75.41}    &35.63 &  68.52     & 36.78 & 68.55\\
\hline
Average            &  33.65  & 64.30    &  33.18  &  67.82  &  \textbf{34.00}   & 72.17   & 32.98 & \textbf{76.49}      &33.00   &  72.17    & 33.28 & 71.38\\
\end{tabular}
} 
\end{center}
\end{small}
\caption{ROUGE-L and NLI scores per language on the XLSum test set for Best-ROUGE and Best-NLI checkpoints. Highest scores in each row are in \textbf{bold}.}
\label{tab:test_results_nli_rouge}
\end{table*}

\begin{table*}[ht!]
\begin{small}
\begin{center}
\resizebox{1.75\columnwidth}{!}{
\rowcolors{1}{}{lightgray}
\begin{tabular}{l|c|l|c|l|c|l|c|l}

\rowcolor{white}
\multicolumn{1}{l|}{}&\multicolumn{2}{c|}{\textbf{Filtered}} & \multicolumn{2}{c|}{\textbf{XLSum mT5-Base}} &\multicolumn{2}{c|}{\textbf{Self-ROUGE}} & \multicolumn{2}{c}{\textbf{Random}}  \\

\textbf{Language}  & \multicolumn{1}{c|}{\textbf{ROUGE}} & \multicolumn{1}{c|}{\textbf{NLI}} & \multicolumn{1}{c|}{\textbf{ROUGE}} &  \multicolumn{1}{c|}{\textbf{NLI}} & \multicolumn{1}{c|}{\textbf{ROUGE}} &  \multicolumn{1}{c|}{\textbf{NLI}} & \multicolumn{1}{c|}{\textbf{ROUGE}} &  \multicolumn{1}{c}{\textbf{NLI}} \\

\hline
Amharic               & 35.15 & 74.65 & 31.14 & 55.07 & 33.76 & 74.96 & 34.33 & 76.97 \\
Arabic                & 33.33 & 72.83 & 32.64 & 53.90 & 34.35 & 67.29 & 34.19 & 69.55 \\
Azerbaijani           & 29.28 & 68.03 & 27.14 & 48.94 & 28.04 & 61.06 & 26.83 & 66.14 \\
Bengali               & 34.17 & 76.53 & 30.90 & 61.05 & 32.75 & 73.34 & 34.04 & 75.61 \\
Burmese               & 40.58 & 73.68 & 36.64 & 51.81 & 38.63 & 60.56 & 37.87 & 73.03 \\
Chinese (simp.)       & 35.66 & 76.95 & 37.86 & 61.61 & 36.22 & 70.42 & 36.36 & 76.37 \\
Chinese (trad.)       & 36.49 & 78.15 & 37.78 & 62.45 & 36.70 & 71.09 & 36.66 & 77.58 \\
English               & 33.23 & 80.32 & 34.16 & 62.83 & 37.21 & 75.68 & 37.46 & 77.25 \\
French                & 34.49 & 72.94 & 32.48 & 55.52 & 33.84 & 65.92 & 33.86 & 68.38 \\
Gujarati              & 32.59 & 63.54 & 29.84 & 51.81 & 32.08 & 64.33 & 31.39 & 65.60 \\
Hausa                 & 37.30 & 67.47 & 35.16 & 48.26 & 35.64 & 59.22 & 36.38 & 59.69 \\
Hindi                 & 37.45 & 74.20 & 35.42 & 54.61 & 38.14 & 68.47 & 37.82 & 71.51 \\
Igbo                  & 35.44 & 64.36 & 32.92 & 42.32 & 33.55 & 62.96 & 34.42 & 56.86 \\
Indonesian            & 35.26 & 79.06 & 34.22 & 60.99 & 35.99 & 71.20 & 36.53 & 73.66 \\
Japanese              & 41.30 & 78.60 & 38.81 & 55.37 & 40.41 & 78.43 & 41.00 & 78.67 \\
Kirundi               & 33.63 & 70.66 & 30.12 & 49.06 & 32.27 & 63.90 & 32.32 & 67.49 \\
Korean                & 39.83 & 78.51 & 36.86 & 62.19 & 38.98 & 74.68 & 38.67 & 79.27 \\
Kyrgyz                & 26.54 & 67.99 & 24.95 & 54.10 & 24.49 & 69.23 & 25.36 & 75.01 \\
Marathi               & 31.31 & 67.86 & 29.28 & 49.31 & 27.30 & 66.20 & 27.06 & 66.32 \\
Nepali                & 38.67 & 80.57 & 35.11 & 62.41 & 37.62 & 78.49 & 38.66 & 78.44 \\
Oromo                 & 29.61 & 76.35 & 26.97 & 58.98 & 28.19 & 78.00 & 28.95 & 71.41 \\
Pashto                & 39.88 & 76.34 & 36.91 & 55.39 & 39.35 & 72.90 & 38.89 & 74.32 \\
Persian               & 37.86 & 79.22 & 36.77 & 62.97 & 37.97 & 75.01 & 38.50 & 77.54 \\
Pidgin                & 34.41 & 72.03 & 33.68 & 54.00 & 33.52 & 69.38 & 33.63 & 65.81 \\
Portuguese            & 33.90 & 64.24 & 32.79 & 40.26 & 34.40 & 52.93 & 35.63 & 57.67 \\
Punjabi               & 36.39 & 64.61 & 32.56 & 48.98 & 35.16 & 60.76 & 35.88 & 63.86 \\
Russian               & 28.71 & 69.20 & 28.28 & 52.04 & 29.28 & 65.32 & 30.42 & 65.05 \\
Scottish Gaelic       & 32.95 & 68.14 & 28.55 & 43.12 & 31.83 & 61.42 & 30.92 & 58.67 \\
Serbian (Cyrillic)    & 29.20 & 62.97 & 26.72 & 44.89 & 27.10 & 62.19 & 28.33 & 62.36 \\
Serbian (Latin)       & 28.21 & 59.93 & 24.85 & 38.66 & 25.63 & 58.28 & 27.14 & 58.69 \\
Sinhala               & 36.04 & 74.76 & 31.42 & 57.27 & 31.52 & 68.35 & 34.36 & 76.12 \\
Somali                & 31.35 & 69.42 & 29.10 & 51.16 & 30.71 & 62.08 & 30.17 & 63.61 \\
Spanish               & 27.33 & 62.34 & 26.90 & 43.45 & 27.44 & 55.76 & 27.53 & 59.26 \\
Swahili               & 36.46 & 73.25 & 33.68 & 54.17 & 34.98 & 69.04 & 34.38 & 66.90 \\
Tamil                 & 33.26 & 81.91 & 31.37 & 63.84 & 31.68 & 77.17 & 31.25 & 80.79 \\
Telugu                & 29.11 & 67.83 & 26.39 & 55.12 & 27.54 & 60.45 & 27.82 & 68.07 \\
Thai                  & 30.66 & 73.27 & 28.43 & 49.35 & 29.62 & 63.97 & 29.80 & 71.96 \\
Tigrinya              & 36.37 & 76.71 & 32.24 & 59.20 & 34.69 & 69.19 & 35.23 & 73.26 \\
Turkish               & 33.65 & 74.66 & 32.81 & 51.91 & 33.01 & 66.81 & 33.58 & 68.67 \\
Ukrainian             & 28.99 & 70.87 & 28.32 & 54.65 & 29.08 & 64.65 & 29.64 & 65.28 \\
Urdu                  & 38.18 & 74.81 & 36.64 & 54.70 & 38.65 & 70.80 & 38.67 & 72.61 \\
Uzbek                 & 28.06 & 70.62 & 25.96 & 52.32 & 25.40 & 68.34 & 26.15 & 72.99 \\
Vietnamese            & 36.69 & 73.01 & 32.77 & 53.04 & 36.07 & 68.67 & 37.41 & 69.45 \\
Welsh                 & 33.89 & 76.06 & 31.97 & 49.66 & 34.43 & 66.18 & 33.93 & 67.17 \\
Yoruba                & 37.01 & 68.48 & 33.90 & 50.81 & 35.03 & 67.52 & 35.28 & 68.20 \\ \hline
Average               & 34.00 & 72.17 & 31.85 & 53.41 & 33.12 & 67.39 & 33.44 & 69.62 \\
\end{tabular}
} 
\end{center}
\end{small}
\caption{ROUGE-L and NLI scores per language on the XLSum test set for
  our Filtered model vs. comparison systems. For simplicity, all
  models are compared using their Best-ROUGE checkpoints. XLSum
  mT5-base predictions are taken from the original XLSum paper
  \cite{hasan-etal-2021-xl}. However, we report on the recomputed
  ROUGE-L using the SentencePiece tokenization of mT5 to make it
  comparable with others. See Section~\ref{subsec:comparisons} for
  more details on Self-Rouge and Random baselines.}
\label{tab:test_results_additional baselines}
\end{table*}

\begin{table*}[t!]
\begin{small}
\begin{center}
\resizebox{0.98\columnwidth}{!}{
\rowcolors{1}{}{lightgray}
\begin{tabular}{lrrr}
\multicolumn{1}{l}{\textbf{Language}}  & \multicolumn{1}{r}{\textbf{Family}} &\multicolumn{1}{r}{\textbf{XNLI}} & \multicolumn{1}{r}{\textbf{Resource}}\\
\hline
Amharic & Semitic & non-xnli & Low\\
Arabic & Semitic & xnli & High\\
Azerbaijani & Turkic & non-xnli & Medium\\
Bengali & Indo-European & non-xnli & Medium \\
Burmese & Sino-Tibetan & non-xnli & Low \\
Chinese Simplified & Sino-Tibetan & xnli & High \\
Chinese Traditional & Sino-Tibetan & xnli & High \\
English & Indo-European & xnli & Very High \\
French & Romance & xnli & Medium \\
Gujarati & Indo-European & non-xnli & Medium \\
Hausa & Afro-Asiatic & non-xnli & Medium \\
Hindi & Indo-European & xnli & Very High \\
Igbo & Niger-Congo & non-xnli & Low \\
Indonesian & Austronesian & non-xnli & High \\
Japanese & Japonic & non-xnli & Medium \\
Kirundi & Bantu & non-xnli & Low \\
Korean & Koreanic & non-xnli & Low \\
Kyrgyz & Turkic & non-xnli & Low \\
Marathi & Indo-Aryan & non-xnli & High \\
Nepali & Indo-Aryan & non-xnli & Low \\
Oromo & Afro-Asiatic & non-xnli & Medium \\
Pashto & Indo-Iranian & non-xnli & High \\
Persian & Indo-Iranian & non-xnli & High \\
Pidgin & Unknown & non-xnli & Medium \\
Portuguese & Romance & non-xnli & High \\
Punjabi & Indo-Iranian & non-xnli & Medium \\
Russian & Indo-European & xnli & High \\
Scottish Gaelic & Celtic & non-xnli & Low \\
Serbian Cyrillic  & Indo-European & non-xnli & Medium \\
Serbian Latin & Indo-European & non-xnli & Medium \\
Sinhala & Indo-European & non-xnli & Low \\
Somali & Afro-Asiatic & non-xnli & Low \\
Spanish & Romance & xnli & High \\
Swahili & Bantu & xnli & Medium \\
Tamil & Dravidian & non-xnli & High \\
Telugu & Dravidian & non-xnli & High \\
Thai & Kra-Dai Languages & xnli & Medium \\
Tigrinya & Semitic & non-xnli & Low \\
Turkish & Turkic & xnli & High \\
Ukrainian & Slavic & non-xnli & High \\
Urdu & Indo-european & xnli & High \\
Uzbek & Turkic & non-xnli & Low \\
Vietnamese & Austroasiatic & xnli & High \\
Welsh & Celtic & non-xnli & Medium \\
Yoruba & Niger-Congo & non-xnli & Medium \\

\end{tabular}
} 
\end{center}
\end{small}
\caption{Classification of XLSum languages into families and their
  membership in XNLI.}
\label{tab:language-family}
\vspace{-10px}
\end{table*}

\begin{table*}[t!]
\begin{small}
\begin{center}
\resizebox{1.98\columnwidth}{!}{
\rowcolors{1}{}{lightgray}

\begin{tabular}{l|c|c|c|c|c|c|c|c|c|c|c|c}
\rowcolor{white}
\multicolumn{1}{l|}{}&\multicolumn{4}{c|}{\textbf{Vanilla}} & \multicolumn{4}{c|}{\textbf{Filtered}} &\multicolumn{4}{c}{\textbf{Controlled}} \\

\rowcolor{white}
\multicolumn{1}{l|}{}&\multicolumn{2}{c|}{\textbf{Best-ROUGE}} & \multicolumn{2}{c|}{\textbf{Best-NLI}} &\multicolumn{2}{c|}{\textbf{Best-ROUGE}} & \multicolumn{2}{c|}{\textbf{Best-NLI}} &\multicolumn{2}{c|}{\textbf{Best-ROUGE}} & \multicolumn{2}{c}{\textbf{Best-NLI}}  \\

\textbf{Language}  & \multicolumn{1}{l|}{\textbf{ROUGE}} & \multicolumn{1}{l|}{\textbf{NLI}} & \multicolumn{1}{l|}{\textbf{ROUGE}} &  \multicolumn{1}{l|}{\textbf{NLI}} & \multicolumn{1}{l|}{\textbf{ROUGE}} &  \multicolumn{1}{l|}{\textbf{NLI}} & \multicolumn{1}{l|}{\textbf{ROUGE}} &  \multicolumn{1}{l|}{\textbf{NLI}} & \multicolumn{1}{l|}{\textbf{ROUGE}} &  \multicolumn{1}{l|}{\textbf{NLI}} & \multicolumn{1}{l|}{\textbf{ROUGE}} & \multicolumn{1}{l}{\textbf{NLI}} \\

\hline
English            & 27.94  &  55.52  & 28.52  & 70.70  & 28.69   & 72.83  & 28.42  & \textbf{80.48} & \textbf{28.79} & 69.61 & 28.49 & 74.58 \\ \hline 
\multicolumn{13}{c}{\textbf{Varying Number of Training Resource}} \\ \hline
High & 30.33 & 48.86 & 30.53 & 55.10 & \textbf{31.50} & 63.90 & 30.87 & \textbf{66.72} & 30.54
 & 57.81 & 30.64 & 61.39\\
Medium &31.51 & 53.67 & 30.75 & 56.36 & \textbf{31.67} & 63.96 & 31.03 & \textbf{67.03} & 31.31
& 58.10 & 30.29 & 59.94\\
Low & 33.22 & 56.09 & 32.20 & 58.64 &\textbf{32.84} & 63.86 & 32.04 & \textbf{70.16} & 32.70 & 57.80 & 32.09 & 61.06 \\ \hline

\multicolumn{13}{c}{\textbf{Language Families}} \\ \hline

Indo-European & 33.13 & 46.37 & 34.92 & 47.77 & 36.67 & 51.84 & 38.00
& \textbf{56.02} & \textbf{39.20} & 50.31 & 36.13 & 51.36\\
Romance & 29.02 & 42.96 & 28.47 & 47.11 & \textbf{29.36} & 55.71 & 28.79
& \textbf{61.39} & 28.69 & 51.73 & 28.61 & 53.27 \\
Turkic & 26.27 & 54.41 & \textbf{26.32} & 54.09 & 26.28 & 61.48 & 26.09
& \textbf{68.71} & 25.79 & 57.62 & 25.85 & 58.97\\
Semitic & 32.78 & 51.89 & 32.52 & 60.70 & \textbf{33.38} & 65.12 & 32.21 & \textbf{70.68}
& 32.36 & 55.68 & 32.79 & 61.88\\
Afro-Asiatic & \textbf{31.63} & 53.43 & 30.05 & 58.37 & 31.17 & 62.55 & 30.30 & \textbf{69.09} & 31.15
& 57.13 & 30.43 & 58.86 \\
\end{tabular}
} 
\end{center}
\end{small}
\caption{ ROUGE-L and NLI scores on XLSum development set for best
  checkpoints averaged across language groups.  For training resources
  we consider three groups with varying numbers of training examples:
  High ([70K–10K]), Medium ([10K–6K]), and Low (less than 6K). For
  language families, the Indo-European cluster represents Bengali,
  Gujarati, Hindi, russian, Serbian (Cyrillic and Latin), and Sinhala;
  the Romance cluster comprises of French, Protuguese, and Spanish;
  the Turkic cluster contains Azerbaijani, Kyrgyz, Turkish, and Uzpek;
  Semitic languages are Amharic, Arabic, and Tigrinya; the
  Afro-Asiatic cluster groups together Hausa, Oromo, and Somali;
  finally, the Indo-Iranian cluster represents Pashto, Persian, and
  Punjabi; we omit clusters with two members and singletons.  We also
  create two subsets depending on whether they appear in the XNLI
  dataset used to train our multilingual NLI model (Available)
   or not (Unavailable). Highest scores are in
  \textbf{bold}.}
\label{tab:results_families}
\end{table*}


\begin{table*}[t!]
\begin{small}
\begin{center}
\resizebox{0.98\columnwidth}{!}{
\rowcolors{1}{}{lightgray}
\begin{tabular}{lcr}
\multicolumn{1}{l}{\textbf{Country of Residence}}  & \multicolumn{1}{c}{\textbf{Total Workers}} &\multicolumn{1}{r}{\textbf{\%}} \\ \hline
Ethiopia & 24 & 6.19 \\
Saudi Arabia & 6 & 1.55 \\
Turkey & 15 & 3.87 \\
Azerbaijan & 12 & 3.09 \\
India & 57 & 14.69 \\
Indonesia & 9 & 2.32 \\
United Kingdom & 14 & 3.61 \\
Argentina & 1 & 0.26 \\
United States & 28 & 7.22 \\
Spain & 6 & 1.55 \\
Pakistan & 26 & 6.70 \\
Czech Republic & 1 & 0.26 \\
France & 6 & 1.55 \\
Nigeria & 33 & 8.51 \\
Japan & 7 & 1.80 \\
South Korea & 6 & 1.55 \\
Kyrgyzstan & 8 & 2.06 \\
Hungary & 1 & 0.26 \\
Myanmar & 7 & 1.80 \\
Nepal & 7 & 1.80 \\
Portugal & 8 & 2.06 \\
Kenya & 16 & 4.12 \\
Burundi & 11 & 2.84 \\
Rwanda & 3 & 0.77 \\
Ukraine & 16 & 4.12 \\
Sri Lanka & 6 & 1.55 \\
Somalia & 5 & 1.29 \\
Serbia & 14 & 3.61 \\
Thailand & 5 & 1.29 \\
Uzbekistan & 9 & 2.32 \\
Vietnam & 8 & 2.06 \\
China & 5 & 1.29 \\
Taiwan & 8 & 2.06 \\ \hline
Total & 388 & 100.00 \\ 
\end{tabular}
} 
\end{center}
\end{small}
\caption{Geographic characteristics of our participants.}
\label{tab:annotator-demo}
\vspace{-10px}
\end{table*}

\begin{table*}[t!]
\begin{small}
\begin{center}
\rowcolors{1}{}{lightgray}

\begin{tabular}{l|c|c|c|c|c|c|c}

\rowcolor{white}
&\multicolumn{2}{c|}{\textbf{Vanilla}} & \multicolumn{2}{c|}{\textbf{Filtered}} &\multicolumn{2}{c|}{\textbf{Controlled}} &  \multicolumn{1}{c}{\textbf{Reference}} \\

\rowcolor{white}

\multicolumn{1}{l|}{\textbf{Language}}&\multicolumn{1}{c|}{\textbf{Best-ROUGE}}
& \multicolumn{1}{c|}{\textbf{Best-NLI}}
&\multicolumn{1}{c|}{\textbf{Best-ROUGE}} &
\multicolumn{1}{c|}{\textbf{Best-NLI}}
&\multicolumn{1}{c|}{\textbf{Best-ROUGE}} &
\multicolumn{1}{c|}{\textbf{Best-NLI}}  \\

\hline

amharic & 0.74 & 0.77 & 0.75 & 0.7 & 0.75 & 0.78 & 0.75 \\
arabic & 0.89 & 0.91 & 0.92 & 0.96 & 0.93 & 0.95 & 0.97 \\
azerbaijani & 0.69 & 0.67 & 0.68 & 0.7 & 0.64 & 0.68 & 0.68 \\
bengali & 0.92 & 0.92 & 0.93 & 0.94 & 0.94 & 0.94 & 0.94 \\
burmese & 0.94 & 0.91 & 0.93 & 0.93 & 0.91 & 0.91 & 0.94 \\ 
chinese (simp.) & 0.85 & 0.78 & 0.92 & 0.88 & 0.81 & 0.87 & 0.95 \\
chinese (trad.) & 0.83 & 0.81 & 0.86 & 0.83 & 0.82 & 0.86 & 0.94 \\
english & 0.91 & 0.92 & 0.91 & 0.92 & 0.89 & 0.91 & 0.9 \\
french & 0.92 & 0.91 & 0.91 & 0.92 & 0.88 & 0.94 & 0.93 \\
gujarati & 0.88 & 0.89 & 0.85 & 0.9 & 0.86 & 0.9 & 0.87 \\
hausa & 0.98 & 0.97 & 0.97 & 0.97 & 0.96 & 0.95 & 0.99 \\
hindi & 0.78 & 0.79 & 0.82 & 0.83 & 0.74 & 0.79 & 0.87 \\
igbo & 0.86 & 0.88 & 0.85 & 0.89 & 0.87 & 0.88 & 0.88 \\
indonesian & 0.87 & 0.84 & 0.88 & 0.9 & 0.88 & 0.92 & 0.9 \\
japanese & 0.85 & 0.9 & 0.88 & 0.85 & 0.91 & 0.88 & 0.94 \\
kirundi & 0.79 & 0.81 & 0.78 & 0.82 & 0.8 & 0.8 & 0.88 \\
korean & 0.88 & 0.83 & 0.9 & 0.87 & 0.84 & 0.88 & 0.94 \\
kyrgyz & 0.93 & 0.94 & 0.94 & 0.96 & 0.94 & 0.94 & 0.96 \\
marathi & 0.78 & 0.76 & 0.81 & 0.81 & 0.81 & 0.75 & 0.79 \\
nepali & 0.87 & 0.95 & 0.92 & 0.95 & 0.85 & 0.94 & 0.87 \\
oromo & 0.61 & 0.56 & 0.62 & 0.61 & 0.66 & 0.55 & 0.78 \\
pashto & 0.71 & 0.7 & 0.67 & 0.7 & 0.7 & 0.68 & 0.6 \\
persian & 0.81 & 0.83 & 0.86 & 0.85 & 0.78 & 0.84 & 0.84 \\
pidgin & 0.87 & 0.84 & 0.84 & 0.84 & 0.87 & 0.84 & 0.84 \\
portuguese & 0.92 & 0.94 & 0.97 & 0.95 & 0.94 & 0.96 & 0.97 \\
punjabi & 0.71 & 0.71 & 0.69 & 0.74 & 0.75 & 0.71 & 0.67 \\
russian & 0.63 & 0.66 & 0.64 & 0.7 & 0.68 & 0.68 & 0.69 \\
scottish gaelic & 0.84 & 0.83 & 0.83 & 0.84 & 0.81 & 0.84 & 0.85 \\
serbian (cyrillic) & 0.84 & 0.91 & 0.81 & 0.91 & 0.86 & 0.88 & 0.95 \\
serbian (latin) & 0.85 & 0.82 & 0.8 & 0.81 & 0.88 & 0.84 & 0.94 \\
sinhala & 0.95 & 0.93 & 0.97 & 0.96 & 0.93 & 0.94 & 0.99 \\
somali & 0.86 & 0.84 & 0.83 & 0.91 & 0.86 & 0.86 & 0.91 \\
spanish & 0.94 & 0.96 & 0.95 & 0.98 & 0.9 & 0.98 & 0.98 \\
swahili & 0.93 & 0.95 & 0.95 & 0.95 & 0.94 & 0.97 & 0.92 \\
tamil & 0.86 & 0.84 & 0.85 & 0.86 & 0.84 & 0.84 & 0.82 \\
telugu & 0.87 & 0.89 & 0.87 & 0.9 & 0.85 & 0.9 & 0.91 \\
thai & 0.88 & 0.87 & 0.89 & 0.89 & 0.87 & 0.91 & 0.97 \\
tigrinya & 0.88 & 0.84 & 0.87 & 0.88 & 0.86 & 0.84 & 0.96 \\
turkish & 0.86 & 0.85 & 0.88 & 0.91 & 0.86 & 0.86 & 0.95 \\
ukrainian & 0.9 & 0.93 & 0.91 & 0.94 & 0.87 & 0.94 & 0.97 \\
urdu & 0.64 & 0.67 & 0.63 & 0.67 & 0.68 & 0.67 & 0.59 \\
uzbek & 0.89 & 0.83 & 0.85 & 0.88 & 0.83 & 0.87 & 0.81 \\
vietnamese & 0.97 & 0.96 & 0.95 & 0.95 & 0.92 & 0.96 & 0. 96 \\
welsh & 0.88 & 0.86 & 0.91 & 0.87 & 0.89 & 0.9 & 0.96 \\
yoruba & 0.83 & 0.78 & 0.78 & 0.78 & 0.75 & 0.75 & 0.85 \\ \hline
Avgerage & 0.85 & 0.84 &	0.85 &	0.86 &	0.84 &	0.86 &	0.88\\

\end{tabular}
\end{center}
\end{small}
\caption{Mean human judgments for Summary Quality per language on the
  XLSum \emph{test set} for Best-ROUGE and Best-NLI checkpoints. We
  also include judgments for Reference summaries.}
\label{tab:results_human_eval_q1}
\end{table*}

\begin{table*}[t!]
\begin{small}
\begin{center}
\rowcolors{1}{}{lightgray}

\begin{tabular}{l|c|c|c|c|c|c|c}

\rowcolor{white}
&\multicolumn{2}{c|}{\textbf{Vanilla}} & \multicolumn{2}{c|}{\textbf{Filtered}} &\multicolumn{2}{c|}{\textbf{Controlled}} &  \multicolumn{1}{c}{\textbf{Reference}} \\

\rowcolor{white}

\multicolumn{1}{l|}{\textbf{Language}}&\multicolumn{1}{c|}{\textbf{Best-ROUGE}} & \multicolumn{1}{c|}{\textbf{Best-NLI}} &\multicolumn{1}{c|}{\textbf{Best-ROUGE}} & \multicolumn{1}{c|}{\textbf{Best-NLI}} &\multicolumn{1}{c|}{\textbf{Best-ROUGE}} & \multicolumn{1}{c|}{\textbf{Best-NLI}} &  \\

\hline

amharic & 0.47 & 0.53 & 0.53 & 0.5 & 0.55 & 0.54 & 0.39 \\
arabic & 0.2 & 0.29 & 0.36 & 0.37 & 0.34 & 0.34 & 0.14  \\
azerbaijani & 0.39 & 0.44 & 0.46 & 0.46 & 0.41 & 0.41 & 0.28 \\
bengali & 0.79 & 0.8 & 0.8 & 0.84 & 0.79 & 0.79 & 0.61 \\
burmese & 0.32 & 0.39 & 0.3 & 0.4 & 0.33 & 0.34 & 0.26 \\
chinese (simp.) & 0.47 & 0.5 & 0.49 & 0.62 & 0.47 & 0.47 & 0.31 \\
chinese (trad.) & 0.46 & 0.53 & 0.54 & 0.53 & 0.5 & 0.5 & 0.33 \\
english & 0.38 & 0.45 & 0.46 & 0.52 & 0.46 & 0.4 & 0.3 \\
french & 0.55 & 0.53 & 0.58 & 0.65 & 0.64 & 0.57 & 0.3 \\
gujarati & 0.48 & 0.49 & 0.51 & 0.5 & 0.51 & 0.47 & 0.45 \\
hausa & 0.29 & 0.32 & 0.32 & 0.35 & 0.31 & 0.35 & 0.15 \\
hindi & 0.44 & 0.5 & 0.49 & 0.54 & 0.48 & 0.49 & 0.37 \\
igbo & 0.41 & 0.47 & 0.4 & 0.47 & 0.36 & 0.44 & 0.22 \\
indonesian & 0.38 & 0.42 & 0.43 & 0.49 & 0.46 & 0.43 & 0.14 \\
japanese & 0.11 & 0.16 & 0.13 & 0.19 & 0.16 & 0.16 & 0.07 \\
kirundi & 0.31 & 0.28 & 0.31 & 0.41 & 0.36 & 0.33 & 0.23 \\
korean & 0.43 & 0.51 & 0.48 & 0.58 & 0.49 & 0.44 & 0.2 \\
kyrgyz & 0.59 & 0.63 & 0.63 & 0.67 & 0.62 & 0.67 & 0.34 \\
marathi & 0.61 & 0.65 & 0.7 & 0.72 & 0.68 & 0.64 & 0.52 \\
nepali & 0.51 & 0.5 & 0.45 & 0.59 & 0.55 & 0.48 & 0.24 \\
oromo & 0.46 & 0.45 & 0.46 & 0.5 & 0.48 & 0.44 & 0.48 \\
pashto & 0.59 & 0.6 & 0.56 & 0.56 & 0.61 & 0.55 & 0.37 \\
persian & 0.26 & 0.29 & 0.25 & 0.26 & 0.31 & 0.29 & 0.12 \\
pidgin & 0.29 & 0.32 & 0.32 & 0.41 & 0.38 & 0.31 & 0.16 \\
portuguese & 0.29 & 0.35 & 0.3 & 0.42 & 0.3 & 0.35 & 0.17 \\
punjabi & 0.49 & 0.52 & 0.5 & 0.58 & 0.55 & 0.53 & 0.31 \\
russian & 0.36 & 0.4 & 0.38 & 0.47 & 0.47 & 0.44 & 0.28 \\
scottish gaelic & 0.49 & 0.55 & 0.52 & 0.56 & 0.57 & 0.56 & 0.53 \\
serbian (cyrillic) & 0.41 & 0.47 & 0.43 & 0.48 & 0.52 & 0.45 & 0.3 \\
serbian (latin) & 0.36 & 0.33 & 0.33 & 0.44 & 0.46 & 0.36 & 0.28 \\
sinhala & 0.39 & 0.43 & 0.31 & 0.51 & 0.41 & 0.38 & 0.17 \\
somali & 0.5 & 0.56 & 0.58 & 0.63 & 0.52 & 0.56 & 0.41 \\
spanish & 0.42 & 0.54 & 0.52 & 0.61 & 0.56 & 0.51 & 0.35 \\
swahili & 0.52 & 0.6 & 0.58 & 0.66 & 0.64 & 0.59 & 0.39 \\
tamil & 0.55 & 0.55 & 0.54 & 0.61 & 0.57 & 0.55 & 0.24 \\
telugu & 0.33 & 0.36 & 0.35 & 0.37 & 0.38 & 0.36 & 0.26 \\
thai & 0.43 & 0.46 & 0.37 & 0.55 & 0.5 & 0.52 & 0.3 \\
tigrinya & 0.5 & 0.56 & 0.53 & 0.65 & 0.62 & 0.5 & 0.34 \\
turkish & 0.52 & 0.55 & 0.61 & 0.62 & 0.57 & 0.6 & 0.47 \\
ukrainian & 0.54 & 0.56 & 0.58 & 0.67 & 0.6 & 0.57 & 0.39 \\
urdu & 0.56 & 0.61 & 0.52 & 0.61 & 0.6 & 0.6 & 0.41 \\
uzbek & 0.63 & 0.55 & 0.58 & 0.64 & 0.57 & 0.6 & 0.4 \\
vietnamese & 0.44 & 0.43 & 0.44 & 0.54 & 0.47 & 0.45 & 0.21 \\
welsh & 0.36 & 0.3 & 0.41 & 0.44 & 0.37 & 0.41 & 0.25 \\
yoruba & 0.42 & 0.38 & 0.42 & 0.38 & 0.41 & 0.39 & 0.33 \\ \hline
Average & 0.44 & 0.47 & 0.46 & 0.52 & 0.49 & 0.47 & 0.31 \\

\hline

\end{tabular}
\end{center}
\end{small}
\caption{Mean human judgments for Attribution per language on the
  XLSum test set for Best-ROUGE and  Best-NLI  checkpoints. We also
  include judgments for Reference summaries.}
\label{tab:results_human_eval_q2}
\end{table*}

\begin{table*}[t!]
\begin{small}
\begin{center}
\rowcolors{1}{}{lightgray}

\begin{tabular}{l|c|c|c|c|c|c|c}

\rowcolor{white}
&\multicolumn{2}{c|}{\textbf{Vanilla}} & \multicolumn{2}{c|}{\textbf{Filtered}} &\multicolumn{2}{c|}{\textbf{Controlled}} &   \multicolumn{1}{c}{\textbf{Reference}} \\

\rowcolor{white}

\multicolumn{1}{l|}{\textbf{Language}}&\multicolumn{1}{c|}{\textbf{Best-ROUGE}}
& \multicolumn{1}{c|}{\textbf{Best-NLI}}
&\multicolumn{1}{c|}{\textbf{Best-ROUGE}} &
\multicolumn{1}{c|}{\textbf{Best-NLI}}
&\multicolumn{1}{c|}{\textbf{Best-ROUGE}} &
\multicolumn{1}{c|}{\textbf{Best-NLI}}  \\

\hline
amharic & 0.37 & 0.38 & 0.4 & 0.36 & 0.43 & 0.44 & 0.32  \\
arabic & 0.16 & 0.24 & 0.29 & 0.33 & 0.28 & 0.3 & 0.14 \\
azerbaijani & 0.36 & 0.43 & 0.42 & 0.45 & 0.39 & 0.38 & 0.28 \\
bengali & 0.77 & 0.76 & 0.78 & 0.82 & 0.76 & 0.77 & 0.61 \\
burmese & 0.29 & 0.35 & 0.27 & 0.35 & 0.28 & 0.32 & 0.23 \\
chinese (simp.) & 0.25 & 0.24 & 0.27 & 0.34 & 0.24 & 0.27 & 0.19 \\
chinese (trad.) & 0.4 & 0.44 & 0.48 & 0.43 & 0.45 & 0.41 & 0.32 \\
english & 0.33 & 0.39 & 0.41 & 0.45 & 0.38 & 0.33 & 0.27 \\
french & 0.35 & 0.38 & 0.39 & 0.43 & 0.43 & 0.38 & 0.23 \\
gujarati & 0.32 & 0.34 & 0.33 & 0.33 & 0.35 & 0.36 & 0.28 \\
hausa & 0.28 & 0.32 & 0.29 & 0.32 & 0.3 & 0.32 & 0.15 \\
hindi & 0.39 & 0.45 & 0.42 & 0.51 & 0.42 & 0.44 & 0.34 \\
igbo & 0.35 & 0.4 & 0.34 & 0.42 & 0.32 & 0.38 & 0.19 \\
indonesian & 0.33 & 0.37 & 0.4 & 0.45 & 0.39 & 0.4 & 0.13 \\
japanese & 0.11 & 0.13 & 0.11 & 0.13 & 0.12 & 0.11 & 0.07 \\
kirundi & 0.27 & 0.25 & 0.26 & 0.35 & 0.34 & 0.29 & 0.21 \\ 
korean & 0.34 & 0.36 & 0.39 & 0.42 & 0.36 & 0.33 & 0.17 \\
kyrgyz & 0.53 & 0.59 & 0.6 & 0.63 & 0.59 & 0.64 & 0.31 \\
marathi & 0.57 & 0.59 & 0.65 & 0.66 & 0.64 & 0.57 & 0.5 \\
nepali & 0.35 & 0.37 & 0.35 & 0.47 & 0.39 & 0.37 & 0.17 \\
oromo & 0.28 & 0.26 & 0.27 & 0.33 & 0.35 & 0.23 & 0.36 \\
pashto & 0.39 & 0.41 & 0.38 & 0.4 & 0.37 & 0.39 & 0.26 \\
persian & 0.29 & 0.31 & 0.29 & 0.29 & 0.29 & 0.33 & 0.15 \\
pidgin & 0.26 & 0.29 & 0.29 & 0.37 & 0.35 & 0.28 & 0.16 \\
portuguese & 0.28 & 0.33 & 0.3 & 0.4 & 0.26 & 0.33 & 0.17 \\
punjabi & 0.48 & 0.51 & 0.49 & 0.56 & 0.52 & 0.52 & 0.31 \\
russian & 0.32 & 0.38 & 0.34 & 0.43 & 0.39 & 0.4 & 0.27 \\
scottish gaelic & 0.41 & 0.46 & 0.4 & 0.42 & 0.46 & 0.42 & 0.4 \\
serbian (cyrillic) & 0.34 & 0.43 & 0.34 & 0.41 & 0.43 & 0.38 & 0.28 \\
serbian (latin) & 0.31 & 0.28 & 0.23 & 0.36 & 0.41 & 0.3 & 0.26 \\
sinhala & 0.35 & 0.37 & 0.28 & 0.47 & 0.37 & 0.35 & 0.16 \\
somali & 0.46 & 0.52 & 0.54 & 0.61 & 0.51 & 0.51 & 0.42 \\
spanish & 0.25 & 0.34 & 0.32 & 0.39 & 0.33 & 0.32 & 0.23 \\
swahili & 0.46 & 0.56 & 0.53 & 0.61 & 0.58 & 0.53 & 0.35 \\
tamil & 0.54 & 0.55 & 0.54 & 0.61 & 0.55 & 0.53 & 0.24 \\
telugu & 0.32 & 0.33 & 0.35 & 0.37 & 0.36 & 0.35 & 0.29 \\
thai & 0.38 & 0.39 & 0.34 & 0.49 & 0.46 & 0.47 & 0.29 \\
tigrinya & 0.44 & 0.46 & 0.46 & 0.56 & 0.54 & 0.43 & 0.32 \\
turkish & 0.49 & 0.52 & 0.55 & 0.57 & 0.54 & 0.57 & 0.46 \\
ukrainian & 0.47 & 0.49 & 0.52 & 0.63 & 0.49 & 0.54 & 0.37 \\
urdu & 0.46 & 0.48 & 0.4 & 0.46 & 0.51 & 0.48 & 0.38 \\
uzbek & 0.56 & 0.48 & 0.49 & 0.58 & 0.47 & 0.53 & 0.34 \\
vietnamese & 0.38 & 0.34 & 0.38 & 0.45 & 0.38 & 0.37 & 0.19 \\
welsh & 0.29 & 0.24 & 0.36 & 0.34 & 0.29 & 0.37 & 0.23 \\
yoruba & 0.36 & 0.28 & 0.35 & 0.29 & 0.33 & 0.31 & 0.27 \\ \hline
Average & 0.37 & 0.40 & 0.39 & 0.45 & 0.41 & 0.40 & 0.27 \\
\end{tabular}
\end{center}
\end{small}
\caption{Mean human judgments for Informativeness per language on the
  XLSum test set for Best-ROUGE and  Best-NLI  checkpoints. We also
  include judgments for Reference summaries.}
\label{tab:results_human_eval_q3}
\end{table*}

\begin{table*}[t!]
\begin{small}
\begin{center}
\resizebox{1.98\columnwidth}{!}{
\rowcolors{1}{}{lightgray}

\begin{tabular}{l|c|c|c|c|c|c|c}

\rowcolor{white}
&\multicolumn{2}{c|}{\textbf{Vanilla}} & \multicolumn{2}{c|}{\textbf{Filtered}} &\multicolumn{2}{c|}{\textbf{Controlled}} &  \multicolumn{1}{c}{\textbf{Reference}}  \\

\rowcolor{white}
\multicolumn{1}{l|}{\textbf{Language}}&\multicolumn{1}{c|}{\textbf{Best-ROUGE}} & \multicolumn{1}{c|}{\textbf{Best-NLI}} &\multicolumn{1}{c|}{\textbf{Best-ROUGE}} & \multicolumn{1}{c|}{\textbf{Best-NLI}} &\multicolumn{1}{c|}{\textbf{Best-ROUGE}} & \multicolumn{1}{c|}{\textbf{Best-NLI}} &  \\

\hline

English & 0.91 & \textbf{0.92} & 0.91 & \textbf{0.92} & 0.89 & 0.91 & 0.9 \\ 
\hline
\rowcolor{white}
\multicolumn{8}{c}{\textbf{Training Resources}} \\ \hline

High & 0.83 & 0.83 & 0.85 & 0.86 & 0.83 & 0.85 & \textbf{0.87} \\
Medium & 0.84 & 0.84 & 0.83 & 0.85 & 0.84 & 0.84 & \textbf{0.88}\\
Low & 0.87 & 0.86 & 0.87 & 0.88 & 0.86 & 0.87 & \textbf{0.89}\\ 
 \hline
 \rowcolor{white}
\multicolumn{8}{c}{\textbf{Language Family}} \\ \hline
Indo-European & 0.81 & 0.82 & 0.81 & 0.84 & 0.82 & 0.83 & \textbf{0.86} \\
Romance & 0.93 & 0.94 & 0.94 & 0.95 & 0.9 & \textbf{0.96} & \textbf{0.96} \\
Turkic & 0.84 & 0.82 & 0.84 & \textbf{0.86} & 0.82 & 0.84 & 0.85  \\
Semitic & 0.83 & 0.84 & 0.84 & 0.85 & 0.85 & 0.86 & \textbf{0.90}  \\
Afro-Asiatic & 0.82 & 0.79 & 0.81 & 0.83 & 0.83 & 0.78 & \textbf{0.89} \\
Indo-Iranian & 0.74 & 0.75 & 0.74 & \textbf{0.76} & 0.74 & 0.74 & 0.70 \\
 \hline
\rowcolor{white}
\multicolumn{8}{c}{\textbf{XNLI Training Data}} \\ \hline
Available & 0.85 & 0.85 & 0.86 & 0.88 & 0.84 & 0.87 & \textbf{0.89} \\
Unavailable & 0.84 & 0.84 & 0.84 & 0.86 & 0.84 & 0.85 & \textbf{0.87} \\
\end{tabular}
} 
\end{center}
\end{small}
\caption{Mean human judgments on Summary Quality for the best
  checkpoints averaged across language groups with 1) varying number
  of training resources, 2) language families and 3) depending on
  whether XNLI data is available. See
  Table~\ref{tab:results_lang_groups:test} for more details about
  different groups. Best results in each row are in \textbf{bold}.}
\label{tab:humaneval_q1_lang_groups}
\end{table*}

\begin{table*}[t!]
\begin{small}
\begin{center}
\resizebox{1.98\columnwidth}{!}{
\rowcolors{1}{}{lightgray}

\begin{tabular}{l|c|c|c|c|c|c|c}

\rowcolor{white}
&\multicolumn{2}{c|}{\textbf{Vanilla}} & \multicolumn{2}{c|}{\textbf{Filtered}} &\multicolumn{2}{c|}{\textbf{Controlled}} &  \multicolumn{1}{c}{\textbf{Reference}}  \\

\rowcolor{white}
\multicolumn{1}{l|}{\textbf{Language}}&\multicolumn{1}{c|}{\textbf{Best-ROUGE}} & \multicolumn{1}{c|}{\textbf{Best-NLI}} &\multicolumn{1}{c|}{\textbf{Best-ROUGE}} & \multicolumn{1}{c|}{\textbf{Best-NLI}} &\multicolumn{1}{c|}{\textbf{Best-ROUGE}} & \multicolumn{1}{c|}{\textbf{Best-NLI}} &  \\

\hline

English  & 0.38 & 0.45 & 0.46 & \textbf{0.52} & 0.46 & 0.4 & 0.3 \\ \hline
\rowcolor{white}
\multicolumn{8}{c}{\textbf{Training Resources}} \\ \hline
High & 0.44 & 0.48 & 0.47 & \textbf{0.53} & 0.49 & 0.48 & 0.30 \\
Medium & 0.42 & 0.44 & 0.44 & \textbf{0.49} & 0.48 & 0.45 & 0.31 \\
Low & 0.46 & 0.49 & 0.47 & \textbf{0.55} & 0.50 & 0.49 & 0.31 \\\hline
\rowcolor{white}
\multicolumn{8}{c}{\textbf{Language Family}} \\ \hline
Indo-European & 0.48 & 0.5 & 0.47 & \textbf{0.55} & 0.53 & 0.5 & 0.36 \\
Romance & 0.42 & 0.47 & 0.47 & \textbf{0.56} & 0.5 & 0.48 & 0.28\\
Turkic & 0.53 & 0.54 & 0.57 & \textbf{0.60} & 0.54 & 0.57 & 0.37\\
Semitic & 0.39 & 0.46 & 0.47 & \textbf{0.51} & 0.50 & 0.46 & 0.29\\
Afro-Asiatic & 0.42 & 0.44 & 0.45 & \textbf{0.49} & 0.43 & 0.45 & 0.35\\
Indo-Iranian & 0.45 & 0.47 & 0.44 & 0.47 & \textbf{0.49} & 0.46 & 0.27\\ \hline
\rowcolor{white}
\multicolumn{8}{c}{\textbf{XNLI Training Data}} \\ \hline
Available & 0.44 & 0.49 & 0.49 & \textbf{0.56} & 0.52 & 0.50 & 0.32 \\
Unavailable & 0.44 & 0.46 & 0.45 & \textbf{0.51} & 0.48 & 0.46 & 0.30 \\
\end{tabular}
} 
\end{center}
\end{small}
\caption{Human evaluation results for Attribution for the best
  checkpoints averaged across language groups with 1) varying number
  of training resources, 2) language families and 3) depending on
  whether XNLI is available. See
  Table~\ref{tab:results_lang_groups:test} for more details about
  different groups. Best results in each row are in \textbf{bold}.}
\label{tab:humaneval_q2_lang_groups}
\end{table*}

\begin{table*}[t!]
\begin{small}
\begin{center}
\resizebox{1.98\columnwidth}{!}{
\rowcolors{1}{}{lightgray}

\begin{tabular}{l|c|c|c|c|c|c|c}

\rowcolor{white}
&\multicolumn{2}{c|}{\textbf{Vanilla}} & \multicolumn{2}{c|}{\textbf{Filtered}} &\multicolumn{2}{c|}{\textbf{Controlled}} &  \multicolumn{1}{c}{\textbf{Reference}}  \\

\rowcolor{white}
\multicolumn{1}{l|}{\textbf{Language}}&\multicolumn{1}{c|}{\textbf{Best-ROUGE}} & \multicolumn{1}{c|}{\textbf{Best-NLI}} &\multicolumn{1}{c|}{\textbf{Best-ROUGE}} & \multicolumn{1}{c|}{\textbf{Best-NLI}} &\multicolumn{1}{c|}{\textbf{Best-ROUGE}} & \multicolumn{1}{c|}{\textbf{Best-NLI}} &  \\

\hline

English & 0.33 & 0.39 & 0.41 & \textbf{0.45} & 0.38 & 0.33 & 0.27 \\ \hline
\rowcolor{white}
\multicolumn{8}{c}{\textbf{Training Resources}} \\ \hline
High & 0.37 & 0.40 & 0.40 & \textbf{0.45} & 0.40 & 0.41 & 0.27 \\
Medium & 0.36 & 0.37 & 0.37 & \textbf{0.41} & 0.40 & 0.38 & 0.28\\
Low & 0.39 & 0.42 & 0.40 & \textbf{0.47} & 0.42 & 0.42 & 0.27 \\ \hline
\rowcolor{white}
\multicolumn{8}{c}{\textbf{Language Family}} \\ \hline
Indo-European & 0.41 & 0.44 & 0.39 & \textbf{0.47} & 0.45 & 0.43 & 0.32 \\
Romance & 0.3 & 0.35 & 0.34 & \textbf{0.41} & 0.34 & 0.34 & 0.21 \\
Turkic & 0.48 & 0.5 & 0.52 & \textbf{0.56} & 0.50 & 0.53 & 0.35 \\
Semitic & 0.32 & 0.36 & 0.39 & \textbf{0.41} & \textbf{0.41} & 0.39 & 0.26 \\
Afro-Asiatic & 0.34 & 0.37 & 0.37 & \textbf{0.42} & 0.39 & 0.35 & 0.31 \\
Indo-Iranian & 0.38 & 0.41 & 0.39 & \textbf{0.42} & 0.40 & 0.41 & 0.24 \\ \hline
\rowcolor{white}
\multicolumn{8}{c}{\textbf{XNLI Training Data}} \\ \hline
Available & 0.36 & 0.40 & 0.39 & \textbf{0.45} & 0.41 & 0.41 & 0.28 \\
Unavailable & 0.38 & 0.39 & 0.39 & \textbf{0.44} & 0.41 & 0.40 & 0.27 \\
\end{tabular}
} 
\end{center}
\end{small}
\caption{Human evaluation results for Informativeness for the best
  checkpoints averaged across language groups with 1) varying number
  of training resources, 2) language families and 3) depending on
  whether XNLI data is available. See
  Table~\ref{tab:results_lang_groups:test} for more details about
  different groups. Best results in each row are in \textbf{bold}.}
\label{tab:humaneval_q3_lang_groups}
\end{table*}

\begin{table*}[t]
  \begin{tabular}{p{15cm}} \hline
    \multicolumn{1}{c}{Document} \\ \hline
    Aucun cas de choléra n'avait été détecté en Algérie depuis 1996,
    tandis que la dernière épidémie d'ampleur remonte à 1986. \\\vspace{.1ex}
    Le
    précédent bilan de l'actuel épisode, communiqué vendredi, faisait
    état de 41 cas avérés, dont un décès, pour 88 cas suspects,
    répartis à Alger -5 malades confirmés- et dans trois wilayas
    (préfectures) alentour.\\\vspace{.1ex}

    Un deuxième patient est décédé du choléra
    dans la wilaya de Blida (50 km au sud d'Alger), où 25 cas sont
    confirmés, a annoncé samedi le ministère. \\\vspace{.1ex}

    Selon lui, un cas a été
    confirmé dans la wilaya de Médéa et un cas suspect recensé dans
    celle d'Aïn Defla, toutes deux jusqu'ici épargnées.\\\vspace{.1ex}

    Situées
    respectivement à 80 et 100 km au sud-ouest d'Alger, ces deux
    régions sont limitrophes de celles de Blida et Tipaza (70 km à
    l'ouest d'Alger), où la grande majorité des malades sont
    recensés.  \\ \hline
\multicolumn{1}{c}{} \\ \hline
    \multicolumn{1}{c}{Reference Summary} \\ \hline
Selon les dernières analyses, "46 cas de choléra ont été confirmés
parmi les 139 cas hospitalisés depuis le 7 août 2018", a fait savoir
le ministère dans un communiqué.  \\
\emph{According to the latest analyses, "46 cases of cholera have been
confirmed among the 139 cases hospitalized since August 7, 2018," the
ministry said in a statement.} 
\\
\hline
\multicolumn{1}{c}{} \\\hline
\multicolumn{1}{c}{Vanilla} \\ \hline
Le ministère de la santé algérien a annoncé samedi que l'épidémie de
choléra a fait plusieurs morts et que le nombre de cas suspects a
atteint les 100..  \\
\emph{The Algerian Ministry of Health announced on Saturday that the cholera epidemic has caused several deaths
and that the number of suspected cases has reached 100..} \\
\hline
\multicolumn{1}{c}{} \\ \hline
\multicolumn{1}{c}{Controlled} \\ \hline
Le ministère de la Santé a confirmé samedi l'augmentation du nombre de
cas de choléra en Algérie, dont deux morts. \\
\emph{The Ministry of Health confirmed on Saturday the increase in the number of cholera cases in Algeria,
including two deaths.} \\
\hline
\multicolumn{1}{c}{} \\ \hline
\multicolumn{1}{c}{Filtered} \\ \hline
Le ministère algérien de la Santé a annoncé samedi que 41 cas de
choléra ont été confirmés et 88 cas suspects, selon le ministère. \\
\emph{The Algerian Ministry of Health announced on Saturday that 41 cases of cholera have been confirmed and 88
suspected cases, according to the ministry.} \\
\hline
  \end{tabular}
  \caption{\label{tab:output:fr1} Input XLSum document in French, accompanied
  by reference summary, and summaries generated by the Vanilla,
  Controlled, and Filtered models, respectively. English translations
  of the summaries are shown in \emph{italics}.}
  \end{table*}

\begin{table*}[t]
\begin{tabular}{p{15cm}} \hline
    \multicolumn{1}{c}{Document} \\ \hline
    En septembre dernier, les essais d'un autre vaccin anti-Ebola
    avaient été lancés, également à Oxford. \\\vspace{.1ex}

    Les derniers essais vont porter sur 72 volontaires dont l’âge
    varie de 18 à 50 ans.\\ \vspace{.1ex}

    Des tests préliminaires sur des singes avaient montré que le
    vaccin, mis au point par Janssen Pharmaceutical Companies, confère
    une immunité contre Ebola. \\\vspace{.1ex}

    Les volontaires d'Oxford sont les premiers humains à se soumettre
    au test de ce vaccin expérimental.\\\vspace{.1ex}

    Dr Matthew Snape, de l'Oxford Vaccine Group, une cellule du
    département de Pédiatrie de l'Université d'Oxford, déclare: "notre
    objectif est d'immuniser tous les participants au bout d'un mois."
    \\\vspace{.1ex} 

    "Il s'agit principalement de chercher à comprendre le profil du
    vaccin en matière de sécurité." \\\vspace{.1ex}

    Au cours de ces essais, les volontaires recevront des doses
    supplémentaires un ou deux mois après l'injection de la dose
    initiale. \\\vspace{.1ex}

    Des essais similaires, à petite échelle, sont en cours aux
    Etats-Unis et dans trois autres pays africains épargnés par
    l'épidémie. \\\vspace{.1ex}

    La dose initiale est destinée à préparer le système immunitaire du sujet, et la deuxième vient doper l'orgacd /home/mlap/google/face/latex/nisme à réagir en cas d'infection.
\\ \hline
\multicolumn{1}{c}{} \\ \hline
    \multicolumn{1}{c}{Reference Summary} \\ \hline
Des chercheurs à l'Université d'Oxford ont commencé l'expérimentation
d'un vaccin contre Ebola sur des volontaires bien portants. 
    \\
\emph{Researchers at the University of Oxford have begun experimenting with a vaccine against Ebola
on healthy volunteers.} 
\\
\hline
\multicolumn{1}{c}{} \\\hline
\multicolumn{1}{c}{Vanilla} \\ \hline
Des essais à grande échelle d'un vaccin contre Ebola sont en cours à
Oxford, dans le sud de l'Angleterre. \\
\emph{Large-scale trials of an Ebola vaccine are underway in
Oxford, southern England.} \\
\hline
\multicolumn{1}{c}{} \\ \hline
\multicolumn{1}{c}{Controlled} \\ \hline
 Les essais d'un vaccin anti-Ebola sont en cours à l'Université d'Oxford.\\
\emph{Trials of an Ebola vaccine are underway at the University of Oxford.} \\
\hline
\multicolumn{1}{c}{} \\ \hline
\multicolumn{1}{c}{Filtered} \\ \hline
 Les scientifiques de l'Université d'Oxford ont lancé les essais d'un vaccin expérimental contre Ebola.\\
\emph{Scientists at the University of Oxford have launched trials of an experimental Ebola vaccine.} \\
\hline
  \end{tabular}
  \caption{\label{tab:output:fr2} Input XLSum document in French, accompanied
  by reference summary, and summaries generated by the Vanilla,
  Controlled, and Filtered models, respectively. English translations
  of the summaries are shown in \emph{italics}.}
  \end{table*}

\begin{table*}[t]
\begin{tabular}{p{15cm}} \hline
    \multicolumn{1}{c}{Document} \\ \hline
    Los ecologistas están acusados de piratería. En la imagen, El portavoz de Greenpreace Internacional, Roman Dolgov.  \\\vspace{.1ex}
    
    Los argentinos Miguel Hernan Pérez Orz y Camila Speziale, junto con la brasileña Ana Paula Alminhana Maciel, se enfrentan a cargos de piratería, lo que puede ser penado con hasta 15 años de cárcel en Rusia.  \\\vspace{.1ex}

    El presidente, Vladimir Putin, señaló que los acusados no son piratas, pero no se opuso a las detenciones. Los miembros de Greenpeace arrestados provienen de 18 países. \\ \vspace{.1ex}

    Un tribunal de Rusia ordenó el jueves extender la detención por dos meses a cinco activistas del grupo de arrestados, que presuntamente intentaron escalar una plataforma petrolera en alta mar en el Ártico ruso.  \\\vspace{.1ex}

    Los ecologistas afectados por la medida provienen de Rusia, Polonia, Canadá y Nueva Zelanda. \\\vspace{.1ex}

    Todos los detenidos esperan ser llamados por el tribunal de la ciudad de Murmansk, en el norte de Rusia, para que cada uno de sus casos sea escuchado. 
    \\\vspace{.1ex} 

   Los miembros de Greenpeace aseguran que organizaron una protesta legal y pacífica. \\\vspace{.1ex}
\\ \hline
\multicolumn{1}{c}{} \\ \hline
    \multicolumn{1}{c}{Reference Summary} \\ \hline
Tres latinoamericanos se encuentran en la lista de los 30 activistas de Greenpeace arrestados en Rusia por protestar la semana pasada contra la exploración de petróleo en aguas árticas.
    \\
\emph{Three Latin Americans are on the list of 30 Greenpeace activists arrested in Russia for protesting last week against oil exploration in Arctic waters.} 
\\
\hline
\multicolumn{1}{c}{} \\\hline
\multicolumn{1}{c}{Vanilla} \\ \hline
Activistas argentinos, brasileños y argentinos, que fueron arrestados en Rusia por protestar en el mar Negro en protesta por el cierre de una plataforma petrolera en el norte de Rusia, están acusados de piratería. \\
\emph{Argentine, Brazilian and Argentinian activists, who were arrested in Russia for protesting in the Black Sea to protest the closure of an oil rig in northern Russia, are accused of piracy.} \\
\hline
\multicolumn{1}{c}{} \\ \hline
\multicolumn{1}{c}{Controlled} \\ \hline
Un tribunal ruso ordenó la detención de cinco miembros del grupo ecologista Greenpeace, acusados de intentar escalar una plataforma petrolera en el Ártico. \\
\emph{A Russian court ordered the arrest of three members of the environmental group Greenpeace, accused of trying to scale an oil platform in the Arctic.} \\
\hline
\multicolumn{1}{c}{} \\ \hline
\multicolumn{1}{c}{Filtered} \\ \hline
Los dos argentinos que fueron arrestados en Rusia por protestar contra el desarrollo de una plataforma petrolera en el Ártico están en riesgo de ser encarcelados por piratería.
\emph{The two Argentines who were arrested in Russia for protesting against the development of an oil rig in the Arctic are at risk of being jailed for piracy.} \\
\hline
  \end{tabular}
  \caption{\label{tab:output:es} Input XLSum document in Spanish, accompanied
  by reference summary, and summaries generated by the Vanilla,
  Controlled, and Filtered models, respectively. English translations
  of the summaries are shown in \emph{italics}.}
  \end{table*}


\end{document}